\documentclass[11pt]{article}
\usepackage[final]{acl}
\usepackage{times}
\usepackage{latexsym}
\usepackage{inconsolata}
\usepackage{tikz}
\usepackage{colortbl,xcolor} 
\usepackage{arydshln} 
\usepackage{amsmath}
\usepackage{amssymb} 
\usepackage{multicol}
\usepackage{multirow}
\usepackage{cleveref}
\usepackage{booktabs}
\usepackage{siunitx}
\usepackage{graphicx}
\usepackage{fontawesome5}
\usepackage{subcaption}
\usepackage[most]{tcolorbox} 

\usetikzlibrary{positioning, arrows, shapes.multipart}
\usepackage[T1]{fontenc}
\usepackage[utf8]{inputenc}
\usepackage{tipa}
\usepackage{xspace}
\usepackage{soul}

\usepackage{enumitem,amssymb}
\newlist{todolist}{itemize}{2}
\setlist[todolist]{label=$\square$}
\usepackage{pifont}
\newcommand{\lvsixty}{W2V2P-LV60\xspace}
\newcommand{\xlsrf}{W2V2P-XLSR53\xspace}
\newcommand{\mipa}{MultiIPA\xspace}
\newcommand{\zipactc}{ZIPA-CTC\xspace}
\newcommand{\zipactcns}{ZIPA-CTC-NS\xspace}
\newcommand{\powsm}{POWSM\xspace}
\newcommand{\powsmctc}{POWSM-CTC\xspace}
\newcommand{\gemini}{Gemini 2.5 Flash\xspace}
\newcommand{\wavlm}{WavLM\xspace}
\newcommand{\whisper}{Whisper\xspace}
\newcommand{\qweni}{Qwen3-Omni-Instruct\xspace}

\newcommand{\ez}{\texttt{DYS-ez}\xspace}
\newcommand{\ua}{\texttt{DYS-ua}\xspace}

\newcommand{\eda}{\texttt{L1-eda}\xspace}

\newcommand{\fl}{\texttt{LID-fl}\xspace}
\newcommand{\geov}{\texttt{GEO-v}\xspace}
\newcommand{\pinv}{\texttt{PI-drc}\xspace}

\newcommand{\ipa}[1]{\textipa{#1}}

\newcommand{\wpad}{\textcolor{white!0}{0}}
\definecolor{colorint}{RGB}{100, 143, 255}
\definecolor{colorext}{RGB}{255, 176, 0}

\newcommand{\bench}{PRiSM\xspace}
\newcommand{\ms}[2]{#1~{\scriptsize$\pm$#2}}

\newtcolorbox{promptbox}[2][]{%
  enhanced,
  breakable,
  colback=black!1,
  colframe=black!35,
  boxrule=0.5pt,
  arc=2pt,
  left=6pt,
  right=6pt,
  top=6pt,
  bottom=6pt,
  fonttitle =\bfseries,
  title ={#2},
  #1
}

\title{
\raisebox{-0.5em}{
    \includegraphics[width=1.3cm]{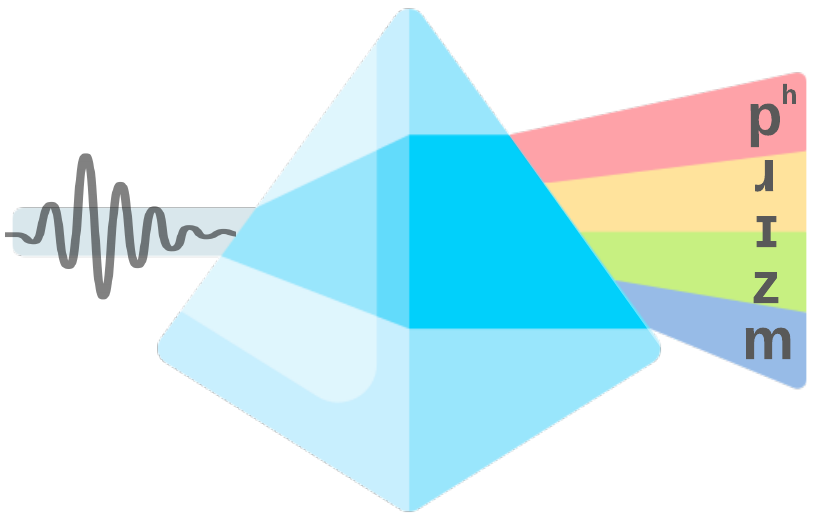}
}\hspace{-0.2em}
PRiSM: Benchmarking Phone Realization in Speech Models
}

\author{
  \textbf{Shikhar Bharadwaj\thanks{Equal contribution.}\textsuperscript{1}} \quad
  \textbf{Chin-Jou Li$^*$\textsuperscript{1}}\quad
  \textbf{Yoonjae Kim$^*$\textsuperscript{1,}\textsuperscript{2}}\quad
  \textbf{Kwanghee Choi\textsuperscript{3}}\quad
  \textbf{Eunjung Yeo\textsuperscript{3}}\\
  \textbf{Ryan Soh-Eun Shim\textsuperscript{4}}\quad
  \textbf{Hanyu Zhou\textsuperscript{1}}\quad
  \textbf{Brendon Boldt\textsuperscript{1}}\quad
  \textbf{Karen Rosero Jacome\textsuperscript{1}}\\
  \textbf{Kalvin Chang\textsuperscript{5}}\quad
  \textbf{Darsh Agrawal\textsuperscript{1}}\quad
  \textbf{Keer Xu\textsuperscript{1}}\quad
  \textbf{Chao-Han Huck Yang\textsuperscript{6}}\\
  \textbf{Jian Zhu\textsuperscript{7}}\quad
  \textbf{Shinji Watanabe\textsuperscript{1}}\quad
  \textbf{David R. Mortensen\textsuperscript{1}}
  \\
  \textsuperscript{1}CMU\quad
  \textsuperscript{2}GIST\quad
  \textsuperscript{3}UT Austin\quad
  \textsuperscript{4}LMU Munich\\
  \textsuperscript{5}UC Berkeley\quad
  \textsuperscript{6}NVIDIA\quad
  \textsuperscript{7}UBC\\
  \texttt{\{sbharad2,chinjoul,dmortens\}@andrew.cmu.edu, rladbswo12@gm.gist.ac.kr}}

\begin{document}
\maketitle

\begin{abstract}
Phone recognition (PR) serves as the atomic interface for language-agnostic modeling for cross-lingual speech processing and phonetic analysis. 
Despite prolonged efforts in developing PR systems, current evaluations only measure surface-level transcription accuracy.
We introduce \bench, the first open-source benchmark designed to expose blind spots in phonetic perception through intrinsic and extrinsic evaluation of PR systems. 
\bench standardizes transcription-based evaluation and assesses downstream utility in clinical, educational, and multilingual settings with transcription and representation probes. 
We find that diverse language exposure during training is key to PR performance, encoder-CTC models are the most stable, and specialized PR models still outperform Large Audio Language Models. 
\bench releases code, recipes, and datasets to move the field toward multilingual speech models with robust phonetic ability\footnote{\url{https://github.com/changelinglab/prism}}.

\end{abstract}

\section{Introduction}

Phone recognition (PR) entails transcribing speech into phonetic units that capture the physical realization of sounds, independent of language-specific phonological constraints.
By preserving acoustic nuances often abstracted away by word- or phoneme-level models\footnote{
For example, \textit{tell} may be transcribed as \textipa{[t\super hE\textltilde]} in Mainstream American English and \textipa{[t\super hEl]} in Scottish English, while 
the phonemic form of \textit{tell} is consistently 
\textipa{/tEl/}.
}, PR provides a robust foundation for cross-lingual speech processing \citep{li2022asr2k, yusuyin2025whistle} and  downstream applications in clinical \citep{shriberg2025clinical, choi-etal-2025-leveraging} and educational settings \citep{tu2018l1pronunciation, inceoglu2023l2assessment}.

\begin{figure}[t]
\centering
  \includegraphics[width=0.92\linewidth]{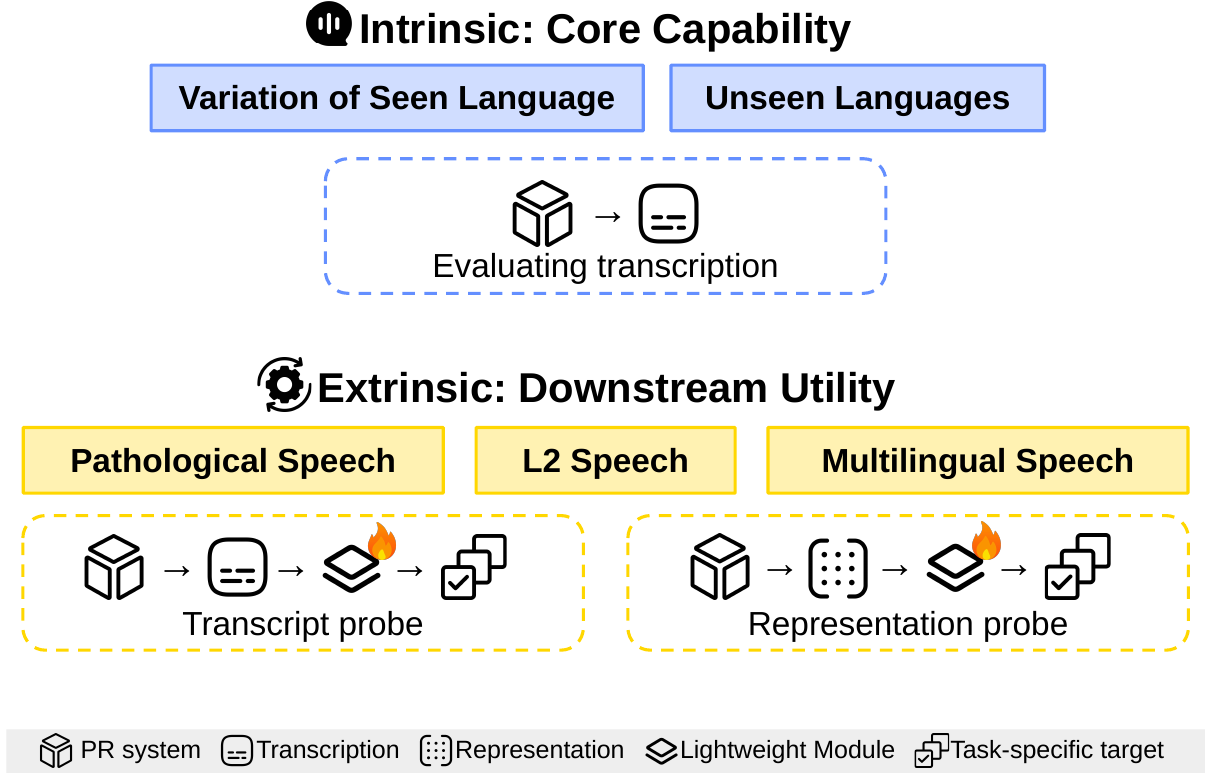}
  \caption{\bench is the first open-source benchmark for phone recognition systems, covering intrinsic and extrinsic evaluations, i.e., transcription task and downstream task performance.}
  \label{fig:phonebench-design}
  \vspace{-1em}
\end{figure}

PR models have scaled substantially to cover diverse linguistic settings (see \autoref{ss:bg-pr-system}), yet existing evaluations remain difficult to compare across studies. 
For example, models often differ in language coverage and phone inventories \cite{zipa}, and evaluation metrics are not standardized \cite{powsm}. 
A common response has been to fix a metric \cite{taguchi23_interspeech, powsm} and expand the number of test datasets to mitigate bias \cite{zipa}. Yet this approach scales poorly due to the scarcity of phonetically transcribed data.
Moreover, transcription error rates do not necessarily reflect a model's phonetic capabilities or practical utility. 
Error rates in PR are inherently noisier than in ASR, as phones, unlike lexical units, correspond to a lower-level, articulatorily defined abstraction of the acoustic signal.

Furthermore, the link between transcription accuracy and downstream performance is often assumed rather than empirically proven. In practice, models leverage phonetic information via two channels: explicit transcriptions and latent internal representations. The latter are especially potent, as they encode rich phonetic cues (see \autoref{ss:bg-phonetic-info}). Consequently, metrics based solely on transcription error fail to capture the full utility and nuanced quality of these representations.

Therefore, we propose \textbf{\bench} to fairly benchmark \textbf{P}hone \textbf{R}ealization \textbf{i}n \textbf{S}peech \textbf{M}odels.
\bench assesses PR systems\footnote{Any pipeline that converts speech into phonetic units.} intrinsically through transcription error, and extrinsically through utility in clinical, educational, and multilingual speech tasks using generated transcriptions and hidden representations. 
\bench applies to PR systems ranging from specialized PR models to general speech-to-text (S2T) systems, including Large Audio Language Models (LALMs), which are increasingly used for general speech tasks despite limited evaluation of their phonetic abilities \citep{peng2024slmsurvey, arora2025landscape}. 

\bench is the first open-source benchmark for PR systems, for which code, evaluation recipes, and datasets are released, where licensing permits.
With its reproducible and expandable framework, \bench supports researchers in understanding model behavior and training strategies, and helps practitioners make informed model choices. 
We evaluate a broad range of PR systems and find that:
(i) \textbf{language exposure matters}: seen languages benefit from familiar patterns, unseen from multilingual training;
(ii) \textbf{data and architecture shape performance}: broad, diverse coverage improves results, while encoder-CTC architectures are more stable;
(iii) \textbf{LALMs lag behind specialized PR models}.

Ultimately, our goal is to establish a common evaluation basis to drive progress toward PR systems that capture robust and generalizable phonetic information across resource conditions.
\begin{table*}[t]
\centering
\small
\renewcommand{\arraystretch}{1.25} 
\setlength{\tabcolsep}{10pt} 

\resizebox{0.95\textwidth}{!}{%
\begin{tabular}{llll}
\toprule
\textbf{Abbr.} & \textbf{Task} & \textbf{Dataset} & \textbf{Lang.} \\
\midrule

\rowcolor{colorint!10}
\multicolumn{4}{l}{\textbf{Intrinsic: Core Capability} (Metrics $\downarrow$ Lower is better)} \\
\addlinespace[0.1cm]

\multicolumn{4}{l}{\hspace{0.3cm}\textit{Phone Recognition (PFER)}} \\
\texttt{PR-tmt} & Variation of Seen Language & TIMIT \cite{garofolo1993timit} & English \\
\texttt{PR-arc} & Variation of Seen Language & L2-ARCTIC Perceived \cite{zhao2018l2arctic} & English \\
\texttt{PR-saa} & Variation of Seen Language & Speech Accent Archive \cite{gmuspeechaccentarchive} & English \\
\texttt{PR-drc} & Unseen Languages & DoReCo \cite{paschen2020doreco} & 45 langs \\
\texttt{PR-vox} & Unseen Languages & VoxAngeles \cite{chodroff2024voxangeles} & 95 langs \\
\texttt{PR-tsm} & Unseen Languages & Tusom2021 \cite{mortensen2021tusom2021} & Tusom \\

\addlinespace[0.1cm]
\midrule
\rowcolor{colorext!10}
\multicolumn{4}{l}{\textbf{Extrinsic: Downstream Utility} (Metrics $\uparrow$ Higher is better)} \\
\addlinespace[0.1cm]

\multicolumn{4}{l}{\hspace{0.3cm}\textit{Pathological Speech: Dysarthria Intelligibility Prediction ($\tau$) \& Child Speech Disorder Detection (F1)}} \\
\texttt{DYS-ez} & Dysarthria Intelligibility Prediction & EasyCall \cite{turrisi21_interspeech} & Italian \\
\texttt{DYS-ua} & Dysarthria Intelligibility Prediction & UASpeech \cite{kim08c_interspeech} & English \\
\texttt{CSD-us} & Child Speech Disorder Detection & UltraSuite \cite{eshky2018ultrasuite} & English \\

\addlinespace[0.2cm]
\multicolumn{4}{l}{\hspace{0.3cm}\textit{L2 Speech: L1 Classification (F1) \& L2 Assessment ($\tau$)}} \\
\texttt{L1-eda} & L1 Classification & EdAcc \cite{sanabria2023edacc} & English \\
\texttt{L1-arc} & L1 Classification & \citet{kominek04_ssw} \& \citet{zhao2018l2arctic} & English\\
\texttt{L2-so} & L2 Assessment & Speechocean762 \cite{zhang2021speechocean762} & English\\

\addlinespace[0.2cm]
\multicolumn{4}{l}{\hspace{0.3cm}\textit{Multilingual: Lang. ID (F1), Geolocation (Recall@1) \& Phone Inventory Induction (F1-PI)}} \\
\texttt{LID-fl} & Lang. ID (LID) & FLEURS-24 \cite{conneau2023fleurs} & 24 langs \\
\texttt{GEO-v}  & Speech Geolocation & Vaani \cite{vaani2025} & Hindi Dialects \\
\texttt{PI-drc} & Phone Inventory Induction & DoReCo \cite{paschen2020doreco} & 45 langs \\

\bottomrule
\end{tabular}
}
\caption{
List of evaluation tasks. \colorbox{colorint!10}{Blue} denotes core capabilities, where lower scores are better. \colorbox{colorext!10}{Yellow} denotes downstream utility, where higher scores are better. \textit{F1-PI} is described in \autoref{appendix:phoneinv}. See \autoref{appendix:datadetails} for license details.
}
\label{tab:evaltasks}
\end{table*}

\section{Background}

\subsection{Phone Recognition Systems}\label{ss:bg-pr-system}
PR can be viewed as a variant of the S2T task that maps speech to phonetic symbols such as IPA \cite{ipa1999handbook}. In this work, we use ``PR system'' to refer broadly to any system capable of converting speech into IPA in a language-agnostic fashion.

Modern PR systems are typically fine-tuned from ASR systems \cite{baevski2020wav2vec, radford2023robust} or trained from scratch on ASR datasets \cite{zhu-etal-2024-taste} with transcriptions automatically converted to IPA using grapheme-to-phoneme (G2P) tools \cite{mortensen2018epitran, zhu2022charsiu-g2p}.
Language-specific approaches \cite{li2020universal, gao21_interspeech} rely on phoneme inventories, while language-agnostic approaches, which we focus on, seek to learn phonetic representations generalized across languages.
LALMs have recently become prominent in speech tasks and have shown competitive performance with cascaded systems that combine LLMs with speech processing modules \cite{yang2025holisticlalm}, motivating interest in their application to PR \cite{huang2025dynamic, wang2025audiobench}.
We describe the systems investigated in this work in \autoref{ss:exp_models}.

\subsection{Phonetic information in PR systems}\label{ss:bg-phonetic-info}
Explicitly generated phonetic transcriptions are easy for humans to inspect and utilize. 
For example, faithful phonetic transcriptions of the speech of a child with a speech sound disorder can help a clinician understand the nature of the disorder and design interventions \cite{dodd2013differential}.
Nevertheless, representing continuous speech with discrete symbols inherently incurs information loss, filtering out non-linguistic variation.

Internal model representations serve as a complement that retains richer information.
Speech models trained on S2T tasks produce temporally aligned representations that capture empirically useful acoustic-phonetic \cite{choi24b_s3mphonetic}, articulatory \cite{cho2024sslartic}, and even semantic \cite{ma-etal-2025-cross} features.
The most widely used S2T model representations are from end-to-end ASR models such as Whisper \cite{radford2023robust} and WavLM \cite{chen2022wavlm}. 
In contrast, LALMs’ representations are often inaccessible or difficult to analyze, as they focus mainly on textual output and lack strict temporal alignment with input speech.

\subsection{Assessing phonetic/phonological ability}
In the text modality, language models are evaluated with text input and output. 
PhonologyBench \cite{suvarna2024phonologybench} evaluates G2P, syllable counting, and rhyme judgment, while \citet{bunzeck2025smalllm} and \citet{goriely2025wordseg} probe phonological knowledge using minimal pairs and word segmentation.
In the speech modality, models are evaluated with speech (and optionally text) input and representation output. 
SUPERB \cite{yang21c_interspeech} and Dynamic-SUPERB \cite{huang2025dynamic} include phoneme recognition, phonological feature analysis, and pronunciation evaluation.
BabySLM \cite{lavechin2023babyslm} and the ZeroSpeech challenges \cite{nguyen2020zerospeech} propose metrics that evaluate phonological and acoustic-phonetic contrasts based on minimal pairs.
In contrast to previous work, \bench evaluates phonetic ability in both text and speech through intrinsic and extrinsic tasks.
\begin{table*}[t]
\centering
\small
\resizebox{\textwidth}{!}{
\begin{tabular}{l l c l l c}
\toprule
\multirow{2}{*}{\textbf{Model}} & \textbf{Architecture} & \multirow{2}{*}{\textbf{Loss}} & \textbf{SSL Pre-training} & \textbf{Phone Recognition} & \multirow{2}{*}{\textbf{Langs}} \\
 & \textbf{(Enc / Dec)} & & \textbf{Data} & \textbf{Data} & \\
\midrule

\lvsixty \cite{xu22b_interspeech} & Enc: Wav2Vec2 & CTC & LibriLight & MLS, CV, Babel & 40+ \\
\xlsrf \cite{xu22b_interspeech} & Enc: XLSR53 & CTC & MLS, CV, Babel & MLS, CV, Babel & 40+ \\
\mipa \cite{taguchi23_interspeech} & Enc: XLSR53 & CTC & MLS, CV, Babel & CV 11.0 & 7 \\

\midrule

\zipactc \cite{zipa} & Enc: Zipformer & CR-CTC & None & IPAPack++ & 88 \\
\zipactcns \cite{zipa} & Enc: Zipformer & CR-CTC & None & IPAPack++ \& PL & 4k \\

\midrule

\multirow{2}{*}{\powsm \cite{powsm}} & Enc: E-Branchformer & \multirow{2}{*}{CTC-Att} & \multirow{2}{*}{None} & \multirow{2}{*}{IPAPack++} & \multirow{2}{*}{88} \\
 & Dec: Transformer & & & & \\

\powsmctc (\href{https://huggingface.co/espnet/powsm_ctc}{ours}) & Enc: E-Branchformer & Int-CTC & None & IPAPack++ & 88 \\

\midrule

\gemini \cite{comanici2025gemini} & Closed & N/A & Closed & Closed & >200 \\

\multirow{2}{*}{\qweni \cite{xu2025qwen3omnitechnicalreport}} & Enc: AuT & \multirow{2}{*}{AR} & \multirow{2}{*}{Closed} & \multirow{2}{*}{Closed} & \multirow{2}{*}{19} \\
 & Dec: MoE Transformer & & & & \\

\bottomrule
\end{tabular}
}
\caption{Included PR systems. 
Architecture abbrv.: Encoder (Enc), Decoder (Dec), Audio Transformer (AuT), Mixture-of-Experts (MoE);
Loss abbrv.: Consistency Regularized CTC (CR-CTC) \cite{crctc}, Hybrid CTC/Attention (CTC-Att) \cite{watanabe2017hybridctc}, Intermediate CTC (Int-CTC) \cite{lee2021intermediate}, Autoregressive (AR);
Data abbrv.: Multilingual LibriSpeech (MLS), Common Voice (CV), Pseudo-labeled (PL).}
\label{tab:model-summary}
\end{table*}

\section{Evaluation Framework of \bench}
\bench covers intrinsic (\autoref{ss:core}) and extrinsic (\autoref{ss:downstream}) evaluations shown in \autoref{fig:phonebench-design}. 
Intrinsic evaluation compares predicted transcriptions to gold labels, while extrinsic evaluation measures transcriptions and internal representations on downstream tasks.
In extrinsic evaluation, transcriptions provide a direct and interpretable signal of explicit phonetic content, whereas representations are commonly used in downstream tasks but may encode non-phonetic information.
\autoref{tab:evaltasks} summarizes included datasets and metrics.

\subsection{Intrinsic: Core Capability}\label{ss:core}
We use \textbf{Phonetic Feature Error Rate (PFER)} to measure the distance between reference and predicted transcriptions. 
Unlike Phone Error Rate (PER), which treats each phone as a token, PFER computes the edit distance $D(\cdot, \cdot)$ over articulatory features $\text{feat}(\cdot)$ such as roundness or voicing \cite{panphon}. As shown in \autoref{eq:pfer}, where $u$ denotes an utterance (sequence of phones where $i$ indexes this sequence) and $u^*$ its ground truth, PFER is calculated as the total feature edit distance across all utterances divided by the total number of phones, representing the percentage of incorrect features. 
\begin{equation}
\label{eq:pfer}\small
    \text{PFER} = \frac{1}{\sum_i |u_i^*|}\sum_i D(\text{feat}(u^*_i),\text{feat}(u_i))
\end{equation}

The tasks comprise two categories: 
\textbf{Variation of seen languages} includes regional and non-native speech, testing whether PR systems rely excessively on seen patterns rather than the actual input.
\textbf{Unseen languages} assess the system’s language-agnostic phonetic knowledge, though for closed LALMs strict verification is not possible because full training corpora are not publicly available.
Details of each task and dataset are in \autoref{appendix:datadetails-pr}.

\subsection{Extrinsic: Downstream Utility}\label{ss:downstream}
We evaluate PR systems using two downstream probes: a \textbf{transcript probe (TP)} and a \textbf{representation probe (RP)}.
TP takes predicted phonetic transcriptions as input and uses a text-based bi-GRU. RP, following the setup in \citet{hear}, uses the final hidden-layer representations with temporal attention pooling and a Multi-Layer Perceptron.
We also explore alternative variants of layer-aggregation in \autoref{appendix:layer}.
Since phonetic transcripts and hidden representations differ substantially in form, TP and RP provide complementary views of downstream phonetic utility, and we therefore focus our comparisons primarily within each probe type.
Metrics for each task are listed in \autoref{tab:evaltasks} and the detailed experimental setup is in \autoref{appendix:setup}.

We consider three categories of downstream tasks where phonetic information is essential. 
In \textbf{pathological speech assessment}, phonetic transcriptions are used to document patients' speech and support diagnosis and treatment planning \cite{ball2009importance, nelson2020use}. 
In \textbf{L2 speech assessment}, phonetic cues enable pronunciation feedback \cite{Franco2010EduSpeakAS} and accent classification \cite{angkititrakul2006phone-based-accent}. 
In \textbf{multilingual speech identification}, analyzing phonetic and phonological differences across languages and dialects, such as phone inventories, phonotactics, and phoneme realization, is crucial \cite{schultz2006multilingualsp}.
We describe each task and dataset in detail in \autoref{appendix:datadetails-downstream}.
\section{Benchmarked Models}
\label{ss:exp_models}

\autoref{tab:model-summary} summarizes the studied model families:
\begin{itemize}[nosep]
\item \textbf{Wav2Vec2Phs}: 
\mipa, \lvsixty, and \xlsrf are fine-tuned variants of Wav2Vec2 \cite{baevski2020wav2vec}, contrastively pre-trained speech SSL models, and differ in pre-training coverage and phone recognition fine-tuning datasets.
\item \textbf{ZIPAs}:
\zipactc and \zipactcns are encoder-CTC models trained from scratch on multilingual data, with \zipactcns further trained on large-scale pseudo-labeled data from \zipactc.
\item \textbf{POWSMs}:
\powsm is an attention-based encoder-decoder (AED) model trained on the same dataset as ZIPAs and augmented for other S2T tasks. Following their framework, we train \powsmctc, an encoder-CTC variant for comparison.
\item \textbf{LALMs}:
We include \gemini (closed-source) and \qweni (open-weight), both state-of-the-art systems widely used in recent studies \cite{lee2025ahelm}. 
Since their representations are difficult to access or pool, we primarily probe them with zero-shot prompting, which is a form of context-based fine-tuning \cite{petrov2023prompting}, and additionally report few-shot results.
The prompts are in \autoref{appendix:prompt}.
\item \textbf{Other baselines}:
We include a naive baseline that randomly predicting the class or the most frequent location (\geov). 
We also include \wavlm\footnote{\url{https://huggingface.co/microsoft/wavlm-base}}\cite{chen2022wavlm} and \whisper\footnote{\url{https://huggingface.co/openai/whisper-small}}\cite{radford2023robust} as competitive baselines for representation probing.
\end{itemize}
\section{Results and Discussion}
\autoref{tab:phonebench:int-results} presents PR performance, and \autoref{tab:phonebench:ext-results} presents a comprehensive breakdown of downstream evaluations.
In general, \zipactcns performs well in all settings, while \whisper excels in RP.
LALMs generally remain less competitive.

\begin{table*}[tb!]
\centering
\resizebox{0.85\textwidth}{!}{%
\begin{tabular}{ll ccc >{\columncolor{colorint!10}}c ccc >{\columncolor{colorint!10}}c}
\toprule
 & & \multicolumn{4}{c}{\textbf{Variation of Seen Language}} &
 \multicolumn{4}{c}{\textbf{Unseen Languages}} \\
\cmidrule(lr){3-6} \cmidrule(lr){7-10}

\textbf{Model} &  &
{\texttt{PR-tmt} } & {\texttt{PR-arc} } & {\texttt{PR-saa} }  & Avg. & {\texttt{PR-drc} } & {\texttt{PR-vox} } & {\texttt{PR-tsm} } & Avg. \\
\midrule

MultiIPA$^*$          & & 16.3 & 15.5 & 13.8 & 15.2 & 18.3 & 15.2 & 30.5 & 21.3 \\
\lvsixty       & & 13.2 & 10.9 & \wpad9.4 & 11.2 & 17.8 & 15.7 & 24.9 & 19.5 \\
\xlsrf         & & 13.5 & \wpad9.9 & \wpad9.0  & 10.8 & 17.3 & \textbf{13.9} & 31.9 & 21.0 \\
\zipactc       & & \textbf{13.1} & \textbf{\wpad9.7} & \wpad9.0 & \textbf{10.6} & 18.0 & 17.0 & 23.7  & 19.6 \\
\zipactcns     & & \textbf{13.1} & \textbf{\wpad9.7} & \textbf{\wpad8.9} & \textbf{10.6} & \textbf{16.8} & 17.1 & 23.1  & 19.0 \\
\powsm         & & 13.7 & 11.3 & 27.6 & 17.5 & 17.1 & 17.1 & \textbf{22.0} & \textbf{18.7} \\
\powsmctc      & & \textbf{13.1} & 10.3 & 10.0 & 11.1 & 18.1 &  15.3 & 32.2 & 21.9 \\
\midrule
Gemini 2.5 Flash$^{**}$    & & 15.2 & 12.7 & 13.2 & 13.7 & 105.3\wpad & 19.7 & 36.3  & 53.8 \\
Qwen3-Omni-Instruct$^{**}$     & & 15.1 & 11.9 & \wpad9.1 & 12.0 & 150.2\wpad & 49.0 & 117.1\wpad  & 105.4\textcolor{colorint!10}{0} \\

\bottomrule
\end{tabular}
}
\vspace{-5pt}
\caption{PFER of the intrinsic evaluation ($\downarrow$). $^*$English is included during pretraining but not fine-tuning. $^{**}$Some of the ``unseen languages'' may have appeared in the training data. See \autoref{sec:prperf} for details.}
\label{tab:phonebench:int-results}
\vspace{-5pt}
\end{table*}
\begin{table*}[tb!]
\centering
\resizebox{0.95\textwidth}{!}{%
\begin{tabular}{lc lll lll llc >{\columncolor{colorext!10}}c}
\toprule
 & & \multicolumn{3}{c}{\textbf{Pathological Speech}} &
 \multicolumn{3}{c}{\textbf{L2 Speech}} &
 \multicolumn{3}{c}{\textbf{Multilingual Speech}} & 
 \multirow{2}{*}{\shortstack{\textbf{{}} \\ \textbf{Score}}} \\
\cmidrule(lr){3-5} \cmidrule(lr){6-8} \cmidrule(lr){9-11}

\textbf{Model} &  &
{\texttt{DYS-ez} } & {\texttt{DYS-ua} } & {\texttt{CSD-us} } & {\texttt{L1-eda} } & {\texttt{L1-arc} } & {\texttt{L2-so} } & {\texttt{LID-fl} } & {\texttt{GEO-v}} & {\texttt{PI-drc}} &  \\
\midrule

Naive Baseline & & \ms{\wpad0.7}{1.6} & \ms{-0.8}{0.9} & \ms{41.8}{1.0} & \ms{\wpad6.3}{0.4} & \ms{14.3}{0.3} & \ms{\wpad1.5}{1.0} & \ms{\wpad4.3}{0.3} & \ms{\wpad3.3}{0.0} & --- & 8.9 \\
\rowcolor{gray!12}
\addlinespace[0.10cm]
\multicolumn{12}{l}{\textbf{Transcript Probe (TP)}} \\
\addlinespace[0.05cm]
\mipa      & & \ms{48.2}{0.2} & \ms{45.6}{1.4} & \ms{93.6}{1.6} & \textbf{\ms{10.0}{1.0}} & \textbf{\ms{50.5}{0.9}} & \ms{33.3}{1.7} & \ms{89.3}{0.5} & \ms{44.5}{0.4} & 40.9 & \textbf{44.3} \\
\lvsixty   & & \ms{42.4}{1.3} & \ms{50.3}{0.9} & \ms{95.6}{1.4} & \ms{\wpad7.6}{0.5}  & \ms{38.0}{0.3} & \ms{36.1}{1.7} & \ms{91.4}{0.2} & \textbf{\ms{45.7}{0.9}} & 51.3 & 42.0 \\
\xlsrf     & & \ms{49.2}{0.8} & \ms{47.6}{0.8} & \ms{92.3}{2.6} & \wpad\underline{\ms{9.1}{0.6}}  & \underline{\ms{43.1}{0.6}} & \underline{\ms{37.5}{0.8}} & \ms{94.1}{0.2} & \ms{44.5}{1.2} & 56.9 & 43.8 \\
\zipactc   & & \underline{\ms{55.0}{0.6}} & \textbf{\ms{57.0}{0.5}} & \ms{91.7}{2.3} & \ms{\wpad6.6}{0.4}  & \ms{30.5}{0.5} & \ms{36.6}{2.8} & \underline{\ms{95.6}{0.2}} & \ms{44.1}{1.0} & 55.2 & 43.5 \\
\zipactcns & & \textbf{\ms{56.6}{0.8}} & \underline{\ms{51.1}{1.3}} & \textbf{\ms{99.4}{0.5}} & \ms{\wpad6.7}{0.3}  & \ms{30.0}{0.3} & \textbf{\ms{40.8}{0.8}} & \textbf{\ms{95.9}{0.1}} & \underline{\ms{44.7}{1.8}} & \underline{56.6} & \underline{44.2}\\
\powsm     & & \ms{52.7}{1.7} & \ms{46.1}{0.8} & \ms{94.3}{1.3} & \ms{\wpad6.5}{0.8}  & \ms{28.0}{0.3} & \ms{28.4}{2.2} & \ms{95.1}{0.5} & \ms{43.7}{1.4} & 48.7 & 39.6 \\
\powsmctc  & & \ms{53.3}{0.4} & \ms{46.5}{0.6} & \ms{96.9}{0.7} & \ms{\wpad6.4}{0.5}  & \ms{29.8}{0.1} & \ms{26.8}{0.7} & \ms{90.4}{0.4} & \ms{42.9}{0.8} & \textbf{57.7} & 40.2 \\
\midrule
\gemini    & & \ms{27.9}{0.6} & \ms{38.5}{0.4} & \ms{95.0}{1.6} & \ms{\wpad6.4}{0.4}  & \ms{22.3}{0.4} & \ms{20.1}{1.1} & \ms{91.8}{0.3} & \ms{33.2}{1.0} & 39.1 & 31.6 \\
\qweni     & & \ms{52.5}{1.8} & \ms{49.4}{1.0} & \underline{\ms{98.9}{0.8}} & \ms{\wpad6.9}{0.6}  & \ms{30.5}{0.3} & \ms{15.6}{1.4} & \ms{89.3}{0.2} & \ms{34.7}{1.2} & 44.5 & 39.2 \\

\addlinespace[0.10cm]
\midrule
\midrule

\rowcolor{gray!12}
\addlinespace[0.10cm]
\multicolumn{12}{l}{\textbf{Representation Probe (RP)}} \\
\addlinespace[0.05cm]
\mipa       & & \ms{65.5}{4.0} & \ms{77.0}{1.4} & \ms{98.5}{0.8} & \ms{11.7}{1.1} & \ms{53.0}{3.3} & \ms{46.3}{1.9} & \ms{78.2}{1.0} & \ms{24.5}{3.7}  & --- & 56.5 \\
\lvsixty    & & \ms{67.2}{1.8} & \underline{\ms{79.9}{0.9}} & \ms{98.6}{0.6} & \ms{12.0}{0.3} & \ms{60.7}{2.9} & \ms{49.9}{1.0} & \ms{76.6}{1.3} & \underline{\ms{24.6}{2.1}}  & --- & 59.4 \\
\xlsrf      & & \ms{70.8}{2.2} & \textbf{\ms{82.0}{2.2}} & \ms{99.2}{0.7} & \ms{13.0}{1.1} & \ms{47.0}{6.0} & \ms{50.7}{3.1} & \ms{81.0}{2.3} & \ms{21.5}{3.1}  & --- & 58.2 \\
\zipactc    & & \ms{73.2}{2.2} & \ms{74.7}{1.2} & \textbf{\ms{99.5}{0.3}} & \ms{13.9}{0.9} & \ms{73.4}{2.5} & \ms{54.0}{0.8} & \ms{96.1}{0.7} & \ms{23.0}{1.2}  & --- & \underline{62.9} \\
\zipactcns  & & \ms{71.2}{2.2} & \ms{75.1}{0.8} & \ms{98.6}{0.9} & \ms{13.7}{0.6} & \underline{\ms{74.1}{2.8}} & \underline{\ms{54.3}{0.5}} & \textbf{\ms{96.8}{0.3}} & \ms{24.0}{1.5}  & --- & 62.7 \\
\powsm      & & \ms{73.0}{3.0} & \ms{70.8}{1.1} & \textbf{\ms{99.5}{0.3}} & \ms{10.3}{1.2} & \ms{68.0}{1.9} & \ms{53.1}{0.3} & \ms{96.5}{0.1} & \ms{21.5}{2.2}  & --- & 60.4 \\
\powsmctc   & & \underline{\ms{73.6}{1.5}} & \ms{66.7}{1.6} & \ms{97.9}{0.9} & \ms{\wpad8.0}{0.7}  & \ms{53.0}{0.5} & \ms{45.7}{3.0} & \ms{75.4}{1.5} & \ms{14.1}{2.6}  & --- & 55.2 \\
\midrule
\wavlm      & & \ms{69.2}{2.0} & \ms{77.5}{1.4} & \ms{99.0}{0.5} & \underline{\ms{14.4}{1.0}} & \ms{58.3}{2.0} & \ms{50.2}{1.4} & \ms{76.2}{3.2} & \ms{23.5}{4.6}  & --- & 59.4 \\
\whisper    & & \textbf{\ms{74.8}{1.1}} & \ms{79.5}{0.3} & \textbf{\ms{99.5}{0.3}} & \textbf{\ms{24.3}{1.6}} & \textbf{\ms{84.3}{3.0}} & \textbf{\ms{57.2}{0.8}} & \underline{\ms{96.3}{0.5}} & \textbf{\ms{35.0}{2.7}}  & --- & \textbf{68.5} \\

\midrule
\addlinespace[0.10cm]
\rowcolor{gray!12}
\addlinespace[0.10cm]
\multicolumn{12}{l}{\textbf{Zero-shot}} \\
\addlinespace[0.05cm]
\gemini       & & 21.4 & 50.4 & 75.3 & 32.7 & 43.9 & 35.8 & 91.5 & \wpad6.5 & --- & 41.5 \\
\qweni        & & 27.0 & 61.7 & 70.9 & 18.2 & 31.8 & 49.8 & 59.1 & \wpad5.3 & --- & 41.5 \\

\bottomrule
\end{tabular}
}
\vspace{-5pt}
\caption{PR system performance on extrinsic tasks ($\uparrow$). Results are reported as mean $\pm$ standard deviation across 5 random seeds where applicable. Best numbers are \textbf{bolded} and second-best \underline{underlined}. See \autoref{sec:extrinsic} for details.
The formula for aggregrated score is in \autoref{appendix:phonebenchscore}.
}
\vspace{-5pt}
\label{tab:phonebench:ext-results}
\end{table*}

\subsection{Intrinsic Evaluation}
\label{sec:prperf}
We observe a consistent trend for language variation: CTC-based models generally outperform LALMs, followed by AED models. 
For \mipa, English appears during pretraining but not finetuning, highlighting the importance of language coverage in PR data. 
On \texttt{PR-saa}, \powsm performs poorly likely due to decoder search on long speech sequences; meanwhile, a text-based G2P model \cite{zhu2022charsiu-g2p} achieves a PFER of 10.2, beating \gemini despite modeling only canonical pronunciations.

For unseen languages, AED and CTC models show comparable performance, whereas LALMs perform poorly overall.
For closed LALMs, however, strict verification of the unseen-language condition is not possible because full training corpora are not publicly available.
Performance also varies across datasets.
On DoReCo, for example, the high LALM averages are driven by a small number of unstable generations, often involving long insertions or repetitions \cite{neuraltextdegen}, which PFER penalizes more heavily than ordinary substitution errors.
Most utterances nevertheless remain below PFER 50 (86\% for \gemini, 91\% for \qweni), while only 1\% and 6\% exceed 100.
\powsm outperforms \powsmctc and exhibits performance comparable to ZIPAs, suggesting that incorporating a degree of language modeling may improve generalization by capturing shared phonological patterns, as further analyzed in \autoref{sec:analysis_noise}.

These trends show that \textbf{variation in seen languages benefits from outputs grounded in known patterns}, whereas \textbf{unseen languages benefit from multilingual training and learned phonological patterns}.


\subsection{Extrinsic Evaluations}
\label{sec:extrinsic}
For transcript probe, ZIPAs and \xlsrf are generally competitive. 
ZIPAs perform well on pathological speech, likely due to their normalized, smaller vocabularies, which approximate broad transcription known to be reliable for speech disorders \cite{shriberg1991reliability}, 
while \xlsrf benefits from diverse pretraining data. 
Multilingual training further improves performance, especially on multilingual tasks.
We discuss \pinv in \autoref{sec:analysis_phoneinv} as an example.
\whisper's strength in representation probe  suggests that large-scale ASR pretraining produces representations that retain phonetic information.
This overall pattern from RP remains consistent when using a learned weighted sum over layers.
We report those results in \autoref{appendix:layer}.

A trade-off of TP and RP emerges among specialized PR models: for example, Wav2Vec2Phs achieve strong TP results on L2 speech but show limited gains on RP, whereas ZIPAs underperform on TP yet excel on RP. 
Task category also influences their relative performance: Pathological speech benefits more from RP, L2 speech falls in the middle, and multilingual tasks tend to favor TP. 
We hypothesize that transcripts act as a structured bottleneck: pathological speech relies on features such as timbre and prosody, whereas multilingual settings benefit less from acoustic detail.
We investigate the behavior of TP on \geov in \autoref{sec:analysis_geo}.

LALMs show task-dependent performance. 
Notably, \qweni achieves competitive TP on pathological speech, but they generally perform poorly in zero-shot settings and underperform on languages other than English (\ez). 
An exception is L2 speech, where the gap is smaller, explored in \autoref{sec:mallm}. 
Additional few-shot results are reported in \autoref{appendix:fewshot}; they improve pathological speech tasks but show mixed effects elsewhere, with no consistent gains as the number of examples increases.

Overall, our results highlight the \textbf{importance of evaluating PR systems with a combination of intrinsic and extrinsic tasks}. 
Intrinsic evaluation alone may not fully capture phonetic capabilities, while extrinsic evaluation reveals that relative performance on TP and RP is task-dependent. 
\textbf{Multilingual pretraining and fine-tuning improve performance} across model families, and \textbf{encoder-CTC based architectures provide more stable PR performance} in new domains. 
In contrast, \textbf{LALMs remain limited} in phone recognition and related tasks.
\section{Analysis}
We conduct several analyses to anchor our observations. 
In \autoref{sec:analysis_noise}, we examine how architectural choices, especially output dependency, affect the balance between phonotactics and acoustics, echoing with evaluation results.
In \autoref{sec:analysis_phoneinv}, we study multilingual generalization and confirm that encoder-only architectures trained with diverse language coverage at all stages perform well for PR. 
In \autoref{sec:analysis_geo}, we analyze TP in detail and show that it effectively captures phone distribution differences across regions. 
Finally, in \autoref{sec:mallm}, we assess zero-shot performance of LALMs on challenging tasks, concluding that they remain insensitive to sociophonetic variation.

\subsection{Phonotactics or the Acoustic Signal}
\label{sec:analysis_noise}

\begin{figure}
    \centering
    \includegraphics[width=\columnwidth]{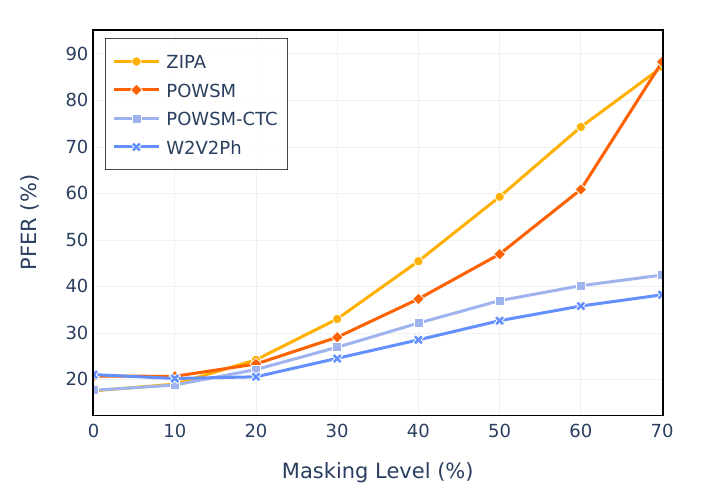}
    \caption{PFER vs Phone masking rate. A PR model that relies only on acoustics should produce a horizontal line. Encoder-only models trained with CTC loss retain acoustic fidelity at high masking levels. See \autoref{sec:analysis_noise}.
    }
    \label{fig:fervsnoise}
    \vspace{-1em}
\end{figure}

Ideally, PR systems would faithfully transcribe the actual pronunciation in the speech signal via acoustic modeling.
Instead, model transcriptions often normalize toward standard pronunciations or other probabilistically likely phone patterns \citep{zipa,powsm}, essentially relying on (phone-level) language modeling \citep{pimentel2020phonotactic}.\footnote{A common example of such phonotactic knowledge is the intuition that \textit{brick} [\textipa{b\*rIk}] is a valid phone sequence in English while \textit{bnick} [\textipa{bnIk}] is not \citep{spe}.}
Additionally, models can also overfit phonotactics from the high-resource languages.
In this experiment, we investigate the extent to which PR systems rely on such phonotactic patterns present in the training data, as opposed to information derived directly from acoustic signal.

Using TIMIT \cite{garofolo1993timit}'s time-aligned phone transcripts, we replace $p\%$ of phones with silence, transcribe the modified speech using PR, and compute PFER against a reference containing only the remaining phones.
In \autoref{fig:fervsnoise}, we plot the phone masking rate against the PFER for different model families.
For a model that only relies on the acoustic waveform for prediction, the curve would be a horizontal line.
However, for models that rely on phonotactics, the PFER will increase with greater noise.
While all models start at a similar PFER, Wav2Vec2Phs and \powsmctc perform better than ZIPAs and \powsm at higher masking levels. 
This pattern is consistent with \textbf{differences in output dependency}: \powsm, as an AED model, conditions each prediction on previously generated outputs, whereas Wav2Vec2Phs and \powsmctc use CTC-style decoding without such dependencies, making them less susceptible to error propagation under masking.

However, ZIPAs are also encoder-only (Zipformer \cite{yao2023zipformer}) models, but they are trained with a consistency regularized CTC (CR-CTC) loss \cite{crctc}.
Their behavior suggests that \textbf{stability is not determined by encoder-only design alone}.
CR-CTC requires a model to produce similar representations despite the noise in input speech.
This explains the behavior observed in our analysis.
We also observe that the insertion rates for different models follow the same trend as the curves in \autoref{fig:fervsnoise}, showing that POWSM and ZIPA produce phonetic transcriptions even when there is no input speech.
Interestingly, \zipactcns and \powsm perform best on unseen languages, but \powsm struggles with seen-language variation, and \zipactcns underperforms on pathological speech. This aligns with the idea that some tasks benefit from learned phonological patterns, while others depend more on capturing acoustic information.

\subsection{Zero-Shot Phone Inventory Induction}
\label{sec:analysis_phoneinv}
Identifying the inventory of phones in a new language is an important linguistic application and often an early step toward developing a standardized transcription system for it.
Such a task requires PR models to recognize phones correctly in unseen phonetic environments.
Therefore, it relies on the phonetic diversity the models have seen in the input speech signal during training.
We explore these behaviors in this set of experiments.

\begin{figure}
    \centering
    \includegraphics[width=\columnwidth]{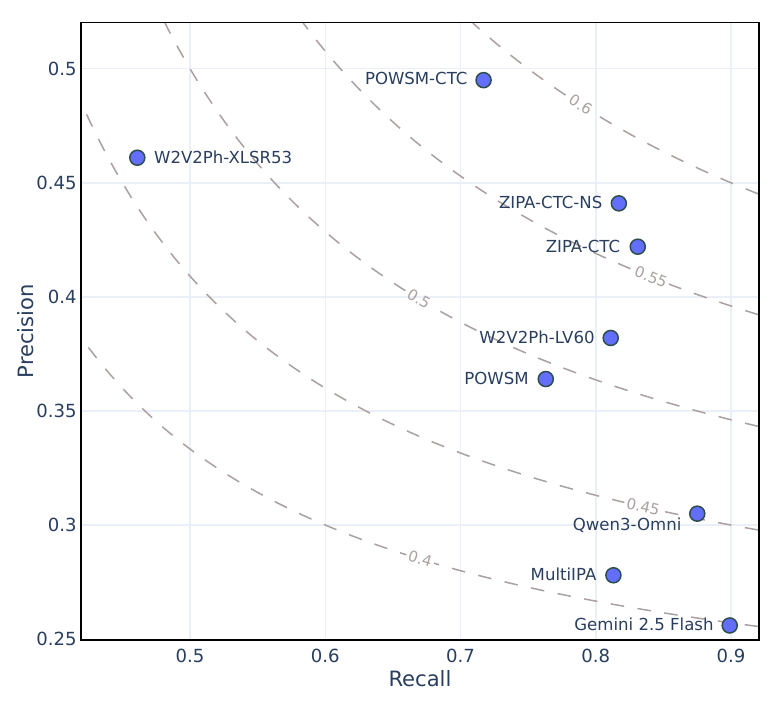}
    \caption{Precision and Recall scores of PR systems on phone inventory induction for unseen languages (\autoref{sec:analysis_phoneinv}). CTC models trained with highly multilingual data are more stable.}
    \label{fig:inv_prec_recall}
    \vspace{-1em}
\end{figure}

Our dataset, derived from DoReCo, consists of low-resource languages absent from the training corpora of all models.
The transcripts from all models are used to compute the phone inventory after applying PanPhon-based phone tokenization \citep{panphon} followed by a set union over detected phones.
The ground truth inventory is constructed similarly using the phonetic transcriptions provided by DoReCo and set similarity metrics (\autoref{appendix:phoneinv}) are computed.
We show the macro-averaged values in \autoref{fig:inv_prec_recall}.

\powsmctc emerges as the strongest model.
The large gap between \powsmctc and \powsm (which differ only in architecture) suggests that the \textbf{encoder-only architecture plays a crucial role in high precision transcripts even in an unseen phonetic environment}.
As for ZIPAs, which differ in training data,  \zipactcns is more precise than \zipactc.
The extended multilingual training of \zipactcns on pseudo-labeled data leads to more precise phone predictions for unseen languages.
This suggests that noisy pseudo-labels allow for improved precision for new languages.
Similar trends is seen in comparing \xlsrf vs \lvsixty and \mipa, where multilingual SSL alone is insufficient for \mipa.
\textbf{Essentially, broader language coverage in both pre-training and supervised training result in a more precise model.}
Although the size of IPAPack++ (17k hr) is much smaller than that used for Wav2Vec2Phs (\textasciitilde160k hr), the larger number of languages in the supervised training stage (88 vs \textasciitilde40) leads to better recall for ZIPAs, compared to Wav2Vec2Phs.
This suggests that \textbf{diversity of languages is as important as the volume of data}.
Most models have a high recall ($>$ 70) and low precision ($<$ 50), suggesting that most predicted phones are incorrect and predictions have a high entropy.

\subsection{Geolocation for Dialectal Speech}
\label{sec:analysis_geo}
In \autoref{tab:phonebench:ext-results}, we observe that our TPs significantly outperform the RPs on Hindi dialectal geolocation \cite{foley-2024-geolocating}, where the former observes an average error of 146 km, while the latter observes an average error of 253 km.
As a reference, our data is spread over 1478 km (East-West) and 1703 km (North-South), covering the entire Hindi speaking region of India.
This performance is surprising, as the cascade-based approach loses suprasegmental information such as intonation that provide strong phonetic cues for the differentiation of dialects \citep{vicenik2013role, grabe2002intonational}. 
However, our results provide empirical evidence that morphological and phonetic differences suffice for fine-grained differentiation between Hindi dialects \citep{Gumperz1958PhonologicalDI}.
\begin{figure}
    \centering
    \includegraphics[width=\columnwidth]{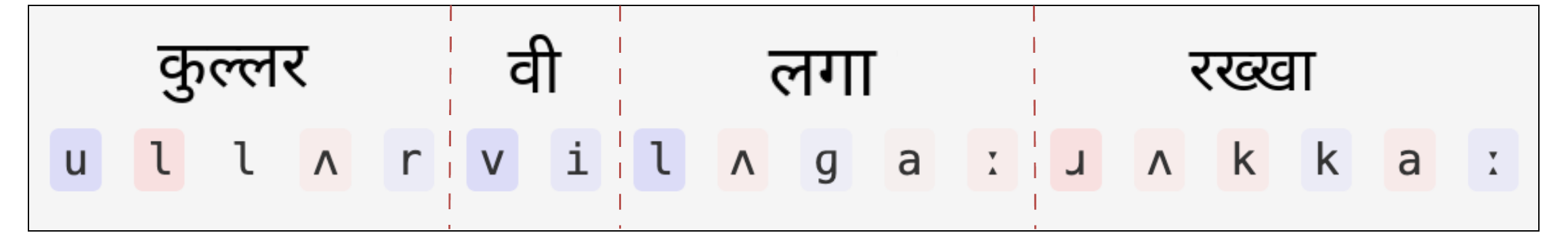}
    \caption{Attribution map from Vaani \citep{vaani2025}. Red supports and blue opposes correct geolocation. \lvsixty detects doubled phones (\autoref{sec:analysis_geo}).}
    \vspace{-1em}
    \label{fig:attr}
\end{figure}

We hypothesize that part of the reason why hidden representations underperform cascade is also due to the downstream probe, where the RP employs attention pooling with an MLP, while the TP employs an RNN.
As the RNN preserves phone order information, even in the case where two dialects share similar phoneme inventories, \textbf{distributional differences of phone sequences between the dialects can be leveraged for fine-grained differentiation} \citep{Gumperz1958PhonologicalDI,shim2024phonotactic}.
We further analyze this behavior by employing integrated gradient based attribution maps \cite{sundararajan2017axiomatic} on TP.
There is a tendency of pronouncing two consonant sounds instead of one in the Bangru dialect of Haryanvi \cite{haryana}.
For example, the English loan word ``Cooler'' \textipa{[ku:lar]} becomes \textipa{[kullar]}, while the Hindi word ``Rakh\textipa{\=a}'' \textipa{[R@.k\super{h}a:]} (kept) becomes \textipa{[R@k.k\super{h}a:]}.
\autoref{fig:attr} shows attribution map for an utterance from \texttt{GEO-v}.
Speaker utters these words in their native accent, \lvsixty outputs \textipa{[ll]} and \textipa{[kk]}, and TP aligns with one of the doubled phones.
We leave a more detailed interpretability analysis to future work.

\subsection{LALMs lack phonetic perception}
\label{sec:mallm}
We examine the zero-shot predictions of LALMs on two tasks: \geov and \eda. 
On \geov, LALMs perform near chance level, whereas on \eda, \gemini achieves the strongest performance.

On \geov, both models exhibit geographic mode collapse. 
\qweni predicts New Delhi for nearly all inputs, while \gemini attains only 6.5\% hit@1 accuracy, with roughly 65\% of its predictions concentrated in 3–4 coordinate clusters near New Delhi (28.6°N, 77.2°E). 
This pattern suggests that \textbf{LALMs have limited sensitivity to dialectal variations} and are strongly biased toward higher-resourced dialects.

Similarly, on \eda, LALMs show a pronounced bias toward the Romance accent cluster, with 25.8\% of Slavic/Balkan and 28.5\% of South Asian accents misclassified as Romance.
Enabling thinking mode exacerbates rather than mitigates such biases by creating more attractor classes.
As a result, the F1-score on \eda drops from 32.7\% to 24.9\%.
Analysis of the reasoning traces reveals that the model over-relies on surface-level phonetic cues, mentioning ``Spanish/Italian/Portuguese'' in 87\% of erroneous Romance predictions and citing ``syllable-timed rhythm'' in 65\% of cases, leading to conflation of phonetically diverse accents.
The confusion matrices for both models are shown in \autoref{appendix:lalm-confusion}.
These findings suggest that LALMs lack the fine-grained acoustic perception, limiting their reliability for tasks requiring unbiased phonetic discrimination.
\section{Conclusion}
We introduce \bench, the first standardized benchmark to measure capabilities of PR systems on transcription task and downstream task performance.
We also open-source our datasets in an easy-to-use format with our toolkit.
Our evaluations reveal that models behave differently on PR and on downstream applications.
Therefore, we recommend that models be benchmarked in both categories to make comparisons.
Our results and analysis show that 
PR for seen language benefits from outputs grounded in familiar patterns, whereas unseen languages' rely on multilingual training and learned phonological patterns. 
Broad and diverse language coverage, along with encoder-CTC architectures, improves stability across tasks, 
while LALMs currently lag behind specialized PR models.
Together, these findings highlight the value of \bench as a framework for evaluating PR systems across diverse languages, tasks, and architectures.

\section*{Limitations}
While \bench evaluates PR systems across a range of intrinsic and extrinsic settings, it is constrained by the availability of curated datasets. As a result, coverage of languages, dialects, accents, and speaking styles remains incomplete and may reflect biases present in the underlying corpora.

In addition, phonetic transcription does not constitute a single objective ground truth: it depends on annotation guidelines, annotator judgments, and the chosen phone inventory. The IPA-based interface may also miss or normalize away language-specific or gradient phonetic phenomena.

Both intrinsic and extrinsic evaluations are necessary to assess PR systems, but each has limitations. Transcript probes align with linguistic features, yet they may also overfit to spurious cues (e.g., sequence length) when datasets are biased or transcripts are noisy due to low PR quality.
For representation probes, phonetic information may be distributed across different layers, and performance can depend on the chosen fusion or pooling strategy.
Models may benefit from task-specific decoding hyperparameters and prompts, whereas we use default settings and prompts that only contain key instructions. 
Our goal is to assess fundamental phonetic capabilities and provide comparative insights; we do not claim that the reported results reflect the best possible performance achievable for each model.

\subsection*{Ethics Statement}
All data used in this work are ethically sourced, either through permissive licensing or with proper consent. 
Speech datasets, particularly those involving pathological speech, may contain sensitive personal information, and we strictly adhere to the licenses and usage conditions associated with each dataset.
PR systems may be misapplied in ways that unfairly label speakers without appropriate expert supervision, especially in educational, clinical, demographic, or geographic contexts.
We introduce \bench with the goal of supporting responsible and rigorous research, and we encourage its use to advance speech technologies that consider linguistic and cultural diversity regardless of resource availability.

\subsection*{The Use of LLMs}
We acknowledge the use of large language models (LLMs) to assist with refinement of the writing, including grammar correction and clarity improvements.
We also used LLMs as coding assistants.
All the code was then verified by authors.
All conceptual, methodological, and experimental work was done independently by the authors.

\section{Acknowledgement}
We thank Jinchuan and Haoran for their support with vLLM, and Brian Yan, Brian Cho and Alexander Metzger for helpful discussions.
This work was supported by National Science Foundation grant \#2504019.
This work was supported by Institute of Information \& communications Technology Planning \& Evaluation (IITP) grant funded by the Korea government(MSIT) (RS-2022-00143911, AI Excellence Global Innovative Leader Education Program).
We also acknowledge the Delta and DeltaAI systems, and support from the NVIDIA Academic Hardware Grant Program 2025.
This work used the Delta and DeltaAI systems at NCSA through allocations CIS210014 and IRI120008P from the Advanced Cyberinfrastructure Coordination Ecosystem: Services \& Support (ACCESS) program.

\bibliography{custom}
\appendix
\section{Dataset details and Licenses}
\label{appendix:datadetails}
This section introduces the datasets and the motivation of downstream tasks.
\autoref{tab:dataset-stats} lists the licensing information and dataset size.

\begin{table}[h]
\resizebox{0.48\textwidth}{!}{%
\begin{tabular}{ll rrr}
\toprule
\textbf{Dataset} & \textbf{Licence} & \textbf{Train} & \textbf{Val\wpad} & \textbf{Test\wpad} \\
\midrule
\multicolumn{3}{l}{\textit{Phone Recognition}}\\
TIMIT & LDC & - & - & 6,300 \\
L2-ARCTIC & CC BY-NC 4.0 & - & - & 3,599 \\
Speech Accent Archive & CC BY-NC-SA 2.0 & - & - & 3,019 \\
DoReCo & CC0 1.0$^*$ & - & - & 18,734 \\
VoxAngeles & CC BY-NC 4.0 & - & - & 5,445 \\
Tusom2021 & MIT & - & - & 2,255 \\
\midrule
\multicolumn{3}{l}{\textit{Pathological Speech}}\\
{{EasyCall}} & CC BY-NC 2.0
 & 11,859 & 4,252 & 4,967 \\
{{UASpeech}} & \href{https://speechtechnology.web.illinois.edu/wp-content/uploads/2024/05/LICENSE.txt}{LICENSE}
 & 9,166 & 5,331 & 6,885 \\
{{UltraSuite}} & CC BY-NC 4.0
& 1,819 & 311 & 287 \\
\midrule
\multicolumn{3}{l}{\textit{L2 Speech}}\\
{{EdAcc}} & CC BY-SA 4.0
& 6,917 & 2,525 & 5,497 \\
{{CMU Arctic}} & \href{http://www.festvox.org/cmu_arctic/cmu_arctic_report.pdf}{Free software}
& 2,264 & 1,132 & 1,132 \\
{{L2-ARCTIC}} & CC BY-NC 4.0
& 13,450 & 6,787 & 6,630 \\
{{SpeechOcean}} & CC BY 4.0
 & 2,260 & 240 & 2,500 \\
\midrule
\multicolumn{3}{l}{\textit{Multilingual Speech}}\\
{{FLEURS-24}} & CC BY 4.0
 & 4,800 & 2,400 & 4,800 \\
{{Vaani-Hi}} & CC BY 4.0
 & 19,780 & 2,668 & 3,985 \\
 DoReCo & CC0 1.0$^*$ & - & - & 18,734 \\
\bottomrule
\end{tabular}
}
\caption{Licence and split size ($\#$utterance). $^*$DoReCo includes datasets of different CC licences; we use the 45-language subset created by \citet{zhu-etal-2024-taste}. See \autoref{appendix:datalinks} for dataset links.}
\label{tab:dataset-stats}
\end{table}

\subsection{Data Links}
\label{appendix:datalinks}
\begin{itemize}
    \item Phone Recognition datasets and EasyCall, EdAcc, CMU-Arctic, L2-ARCTIC, Fleurs-24 and Ultrasuite are available at \url{https://huggingface.co/collections/changelinglab/prism}
    \item UASpeech can be obtained from \url{https://speechtechnology.web.illinois.edu/uaspeech/}
    \item Speechocean762 can be downloaded from \url{https://github.com/jimbozhang/speechocean762}
\end{itemize}

\subsection{Datasets in Intrinsic Evaluation}
\label{appendix:datadetails-pr}
\paragraph{Variation in Seen Language}
TIMIT \citep{garofolo1993timit} contains speech from six regional varieties of American English and is often used for PR evaluation.
The Speech Accent Archive \citep{gmuspeechaccentarchive} provides read speech (the ``Please call Stella'' passage) and narrow phonetic transcriptions from non-native English speakers across 391 L1 languages.
L2-ARCTIC \cite{zhao2018l2arctic} includes read speech from non-native speakers; we use the L2-Arctic Perceived set\footnote{\url{https://huggingface.co/anyspeech}} which consists of manually annotated phoneme transcriptions rather than standard G2P output.

\paragraph{Unseen Languages}
DoReCo \citep{paschen2020doreco} is a dataset of 50+ small or endangered languages with broad phonetic transcriptions; we use the same DoReCo subset as \citet{zipa,zhu-etal-2024-taste}.
VoxAngeles \citep{chodroff2024voxangeles} is a cleaned, 95-language version of the UCLA Phonetics Lab Archive \citep{ucla2009}.
Tusom2021 \citep{mortensen2021tusom2021} is a dataset of speech and narrow phonetic transcriptions of individual words in the low-data Tangkhulic language Tusom. We removed tones as none of the models supports them.

\subsection{Datasets in Extrinsic Evaluation}
\label{appendix:datadetails-downstream}

\paragraph{Pathological Speech Assessment}
\textit{Dysarthria intelligibility prediction} predicts dysarthria severity levels based on phonetic representations. Increasing dysarthria severity is associated with reduced intelligibility, for which impaired phoneme production is a major clinical predictor \cite{xue2023assessing}. Two dysarthric speech datasets are evauated: UASpeech \cite{kim08c_interspeech}, an English corpus with speaker-level intelligibility scores, and EasyCall \cite{turrisi21_interspeech}, an Italian corpus annotated with dysarthria severity ratings. 
\textit{Child speech disorder detection} classifies whether a given utterance is produced by a child with speech disorder, supporting applications in speech therapy and the selection of specialized speech models \cite{rosero2025b, rosero2025a}.
We use acoustic recordings from the Ultrasuite corpus \cite{eshky2018ultrasuite}, with manually corrected transcription-audio mismatches. The curated dataset is released with this paper.

\paragraph{L2 Speech Evaluation}
\textit{Proficiency assessment for L2 Learners} uses phonetic information to automatically assess L2 English proficiency. We use utterances and sentence-level scores on a 0-10 scale from Speechocean762 \cite{zhang2021speechocean762}, an L1 Chinese, L2 English corpus. 
\textit{L1 influence classification} classifies a speaker’s L1 (native language) background, which introduces distinctive articulatory patterns into speech in an L2 language \cite{yang23_interspeech, shi21_icassp}.
We use EdAcc \cite{sanabria2023edacc} for one setup, and the other combines L2-ARCTIC \cite{zhao2018l2arctic} for non-native speech with CMU ARCTIC \cite{kominek04_ssw} for native speech. 

\paragraph{Multilingual Speech Identification}
\textit{Language identification} (LID) predicts the language spoken in an utterance from audio input. We use it as a coarse-grained evaluation of whether phonetic representations can distinguish both seen and unseen languages with the 102 languages in FLEURS \cite{conneau2023fleurs}.
\textit{Speech geolocation identification} predicts the origin of a speaker from an utterance in their native language, drawing on systematic phonetic shifts associated with geography, sociolinguistic variation, and language contact \citep{foley-2024-geolocating}. 
We use data from the Hindi-belt of India from Vaani \cite{vaani2025}. The detailed algorithm for this subset creation is explained in \autoref{appendix:vaani}.
\textit{Phone inventory induction} is the task of inferring the set of phones used by language from speech recordings, which is useful for language documentation and helps identify systematic errors during evaluation.
We use DoReCo (\autoref{appendix:phoneinv}) by deriving phone inventories from gold phone transcriptions and comparing them against the predicted transcriptions for each language.

\section{Metrics}
\subsection{Task Metric: F1 of Phone Inventory (F1-PI)}
\label{appendix:phoneinv}
A phone inventory is the set of all phones used in a language.
F1-PI assesses the degree of overlap between the phones transcribed by a system for a given language and the ground truth phone inventory for that language \citep{zelasko2022discovering}.
For two sets $A$ and $B$, the F1-score is defined as the harmonic mean of $|A-B|/|A|$ and $|B-A|/|B|$.
Set membership can be based on exact matches or fuzzy matches (e.g., over phonetic features).
This metric requires only a reference inventory for the target language, not a full transcription (although inventories can be derived from transcriptions), making it especially useful for under-resourced languages.

\subsection{Summary Metric: \bench Extrinsic Score}
\label{appendix:phonebenchscore}
To aggregate performance across extrinsic evaluation tasks with significantly varying test set sizes ($N_i$) (\autoref{tab:dataset-stats}), we compute a score using a logarithmically weighted average. 
This approach ensures that larger datasets contribute more to the final score due to their statistical significance, while preventing them from completely dominating smaller, high-variance datasets (such as \texttt{CSD-us}).

Let $s_i$ be the model performance on task $i$ and $N_i$ be the number of samples in that task. The aggregate score $S$ is defined as:
\begin{equation}
    S = \frac{\sum_{i=1}^{K} \ln(N_i) \cdot s_i}{\sum_{i=1}^{K} \ln(N_i)}
\end{equation}
where $K=6$ corresponds to the tasks (\texttt{DYS-ez}, \texttt{DYS-ua}, \texttt{CSD-us}, \texttt{L1-eda}, \texttt{L1-arc}, and \texttt{L2-so}) that show differentiation in model behavior.
The weights $w_i = \ln(N_i)$ dampen the linear disparity between the largest ($N=7762$) and smallest ($N=287$) test sets.

We further assess the sensitivity of the aggregate score to the choice of weighting scheme by comparing the resulting model rankings under four alternatives: uniform, $\sqrt{N_i}$, $N_i$, and $1/N_i$, where $N_i$ denotes the number of samples in task $i$. We measure ranking consistency with our $\ln(N_i)$-weighted metric using Kendall's $\tau$.

For TP, the top-3 cluster (\mipa, \zipactcns, and \xlsrf) is preserved under all weighting schemes ($\tau \geq 0.83$), except under the $1/N_i$ scheme, which assigns 77\% of the total weight to the smallest dataset (\texttt{CSD-us}, $N=287$). For RP, rankings are also stable across weighting schemes ($\tau \geq 0.78$), with perfect agreement under $\sqrt{N_i}$ and $N_i$ weighting ($\tau = 1.0$); notably, Whisper remains the top-ranked model under every scheme tested.

\section{Experimental Setup}
\label{appendix:setup}

\paragraph{Probe Details}
All transcript probes use a 2 layer bi-directional GRU with mean pooling to get transcript level representation.
The GRU operates on a character vocabulary built from all predicted transcripts.
GRU has a hidden dimension of 256 and input dimension of 128 with a dropout of 0.1.

For hidden representation probes we use the last layer's hidden representation and attention pool over time to obtain utterance level representation.
This is followed by an MLP composed of 2 linear layers.
First layer's input dimension is the same as the dimension of the model being evaluated.
It outputs an embedding half of this size and the final layer outputs a single scalar for assessment taks, logits over classes for classification tasks, or a unit \texttt{[x y z]} vector for geolocation task.
MSE loss is employed for regression, cross-entropy for classification and angular error loss \cite{foley-2024-geolocating} for geolocation.

\paragraph{Hyper-parameters}
All the experiments can be reproduced via our open-sourced toollkit.
We use a learning rate of 2e-4 for all hidden representation probes and a learning rate of 1e-3 for the cascade probes.
We use the validation F1 (for classification), Kentall Tau (for assessment) and error (in km for geolocation) as early stopping metrics with a patience of 5 epochs and minimum epochs set to 10.
The checkpoint achieving best validation values on these metrics is selected for reporting numbers.

\paragraph{Compute spent}
Each TP probe runs in at most 15 minutes on a single 40GB GPU.
Each RP probe runs in at most 3 hours on a single 40GB GPU.
For TP and RP based extinsic evaluations a total of around 1k GPU hours were spent to get final numbers.
We used almost 1k GPU hours during development phase of the evaluation toolkit as well.
Besides, \bench supports distributed inference that scales to multiple GPUs and supports VLLM \footnote{\url{https://github.com/vllm-project/vllm}}.
For inference, we utilized around ~500 GPU hours including debugging and development costs.
Each \powsmctc model trains on 4 nodes with 4-80GB GPU each and takes 1.5 days to train, amounting to 600 GPU hours for one run.
We can assume another 2k GPU hours for development and experimentation.

\section{L1 to accent cluster mapping for EdAcc}
The EdAcc corpus contains 41 distinct L1 labels, which we consolidate into 13 accent clusters based on phonological and typological similarity.
Grouping criteria include language family (e.g., Sino-Tibetan, Austronesian), vowel inventory size (e.g., 5-vowel Romance languages), prosodic patterns (e.g., syllable-timed vs.\ stress-timed), and shared phonetic transfer patterns to English (e.g., rhoticity, vowel reduction).
\autoref{tab:edacc-l1-accent-mapping} lists the complete mapping.

\label{appendix:edacc-l1-accent-mapping}
\begin{table}[t]
\centering
\scriptsize
\setlength{\tabcolsep}{6pt}
\begin{tabular}{l l}
\toprule
\textbf{EdAcc L1 label} & \textbf{Accent cluster} \\
\midrule
Hindi & SOUTH\_ASIAN \\
Indian English & SOUTH\_ASIAN \\
Urdu & SOUTH\_ASIAN \\
Sinhalese & SOUTH\_ASIAN \\
\midrule
English & INNER\_CIRCLE\_ENGLISH \\
Southern British English & INNER\_CIRCLE\_ENGLISH \\
Mainstream US English & INNER\_CIRCLE\_ENGLISH \\
South African English & INNER\_CIRCLE\_ENGLISH \\
\midrule
Scottish English & CELTIC\_ENGLISH \\
Irish English & CELTIC\_ENGLISH \\
\midrule
Spanish & ROMANCE \\
Spanish (Mexican) & ROMANCE \\
Catalan & ROMANCE \\
Italian & ROMANCE \\
Portoguese & ROMANCE \\
Maltese & ROMANCE \\
\midrule
French & GALLO\_ROMANCE \\
\midrule
Indonesian & INSULAR\_SEA \\
Bahasa & INSULAR\_SEA \\
Filipino & INSULAR\_SEA \\
Tagalog & INSULAR\_SEA \\
\midrule
Vietnamese & MAINLAND\_SEA \\
\midrule
Mandarin & EAST\_ASIAN \\
Japanese & EAST\_ASIAN \\
Korean & EAST\_ASIAN \\
\midrule
Nigerian English & AFRICAN\_ENGLISH \\
Kenyan English & AFRICAN\_ENGLISH \\
Ghanaian English & AFRICAN\_ENGLISH \\
\midrule
Russian & SLAVIC\_BALKAN \\
Polish & SLAVIC\_BALKAN \\
Bulgarian & SLAVIC\_BALKAN \\
Macedonian & SLAVIC\_BALKAN \\
Montenegrin & SLAVIC\_BALKAN \\
Lithuanian & SLAVIC\_BALKAN \\
Romanian & SLAVIC\_BALKAN \\
\midrule
German & GERMANIC \\
Dutch & GERMANIC \\
Icelandic & GERMANIC \\
\midrule
Arabic & AFROASIATIC\_SEMITIC \\
Hebrew & AFROASIATIC\_SEMITIC \\
\midrule
Jamaican English & CARIBBEAN \\
\bottomrule
\end{tabular}
\caption{Mapping from EdAcc L1 labels (41) to 13 accent clusters used in \texttt{L1-eda}.}
\label{tab:edacc-l1-accent-mapping}
\end{table}

\section{Algorithm for Vaani-Hi}
\label{appendix:vaani}

For the \texttt{GEO-v} task, we construct \textbf{Vaani-Hi}, a Hindi-belt subset of the Vaani corpus \citep{vaani2025}, and release it on Hugging Face.

\paragraph{Sampling}
We focus on 12 Hindi-belt states: Chandigarh, Himachal Pradesh, Delhi, Madhya Pradesh, Jharkhand, Uttarakhand, Bihar, Chhattisgarh, Haryana, Rajasthan, Punjab, and Uttar Pradesh.
From each state, we randomly sample up to 4 districts; for each district, we use up to 4 audio shards and take up to 600 utterances per shard (seed 42).

\paragraph{Filtering and Labeling}
We retain only pincodes with more than 450 utterances to ensure sufficient density per location.
Each pincode is mapped to latitude/longitude using a pincode metadata table; we assign mean coordinates per pincode as the geolocation target.

\paragraph{Splitting and Preprocessing}
Splits are created within each pincode (75\%/10\%/15\% train/val/test) to avoid location leakage.
Audio is resampled to 16\,kHz and clipped to a maximum of 20 seconds.

\section{Effect of Layer Selection on RP}
\label{appendix:layer}

Our main representation probe (RP) results use the last layer's hidden representations following \citet{hear}.
To assess sensitivity to layer choice, we repeat RP with a learnable weighted sum across all layers for each model.
\autoref{tab:layersum} reports weighted-layer-sum RP performance, with changes relative to last-layer RP in parentheses, and includes the best last-layer RP result for reference.

Weighted-layer fusion produces gains on some tasks (notably \geov and \ua) but regressions on others, with no model showing consistent improvement across all tasks.
The best RP results under weighted fusion remain comparable to those from the last layer alone.
These results suggest that last-layer representations are a reasonable default, though task-specific layer selection may offer marginal benefits.

\begin{table*}[t]
\centering
\resizebox{0.95\textwidth}{!}{%
\begin{tabular}{lc lll lll ll}
\toprule
 & &
 \multicolumn{3}{c}{\textbf{Pathological Speech}} &
 \multicolumn{3}{c}{\textbf{L2 Speech}} &
 \multicolumn{2}{c}{\textbf{Multilingual Speech}} \\
\cmidrule(lr){3-5}\cmidrule(lr){6-8}\cmidrule(lr){9-10}
\textbf{Model} & & \multicolumn{1}{c}{\texttt{DYS-ez}} & \multicolumn{1}{c}{\texttt{DYS-ua}} & \multicolumn{1}{c}{\texttt{CSD-us}} & \multicolumn{1}{c}{\texttt{L1-eda}} & \multicolumn{1}{c}{\texttt{L1-arc}} & \multicolumn{1}{c}{\texttt{L2-so}} & \multicolumn{1}{c}{\texttt{LID-fl}} & \multicolumn{1}{c}{\texttt{GEO-v}} \\
\midrule
\rowcolor{gray!12}
\addlinespace[0.10cm]
\multicolumn{10}{l}{\textbf{Weighted-layer-sum RP (mean score; change from last-layer RP)}} \\
\addlinespace[0.05cm]
\mipa       & & 66.0 (+0.5) & 82.2 (+5.2) & 99.3 (+0.8) & 11.0 ($-$0.7) & 60.7 (+7.7) & 50.4 (+4.1) & 80.9 (+2.7) & 34.1 (+9.6) \\
\lvsixty    & & 71.3 (+4.1) & 84.4 (+4.5) & 98.9 (+0.3) & 12.7 (+0.7) & 66.0 (+5.3) & 55.5 (+5.6) & 81.7 (+5.1) & 34.9 (+10.3) \\
\xlsrf      & & 74.5 (+3.7) & \textbf{86.8 (+4.8)} & 99.6 (+0.4) & 12.8 ($-$0.2) & 51.6 (+4.6) & \textbf{56.8 (+6.1)} & 79.1 ($-$1.9) & \textbf{35.5 (+14.0)} \\
\zipactc    & & 71.1 ($-$2.1) & 84.9 (+10.2) & \textbf{100.0 (+0.5)} & 14.9 (+1.0) & 68.8 ($-$4.6) & 53.3 ($-$0.7) & 94.0 ($-$2.1) & 34.3 (+11.3) \\
\zipactcns  & & 69.0 ($-$2.2) & 85.0 (+9.9) & 99.7 (+1.1) & 15.4 (+1.7) & 70.3 ($-$3.8) & 54.7 (+0.4) & \textbf{96.4 ($-$0.4)} & 32.8 (+8.8) \\
\powsm      & & 72.6 ($-$0.4) & 82.3 (+11.5) & 99.2 ($-$0.3) & 14.3 (+4.0) & 68.8 (+0.8) & 55.8 (+2.7) & 94.6 ($-$1.9) & 32.0 (+10.5) \\
\powsmctc   & & \textbf{76.0 (+2.4)} & 74.4 (+7.7) & 99.7 (+1.8) & 12.9 (+4.9) & 58.9 (+5.9) & 52.4 (+6.7) & 84.1 (+8.7) & 24.4 (+10.3) \\
\wavlm      & & 67.8 ($-$1.4) & 81.4 (+3.9) & 99.6 (+0.6) & 14.2 ($-$0.2) & 57.1 ($-$1.2) & 53.2 (+3.0) & 69.5 ($-$6.7) & 30.8 (+7.3) \\
\whisper    & & 75.6 (+0.8) & 78.6 ($-$0.9) & \textbf{100.0 (+0.5)} & \textbf{24.9 (+0.6)} & \textbf{81.6 ($-$3.2)} & 53.9 ($-$3.3) & 92.5 ($-$3.8) & 24.0 ($-$11.0) \\
\midrule
\textbf{Best last-layer RP} & & 74.8 & 82.0 & 99.5 & 24.3 & 84.3 & 57.2 & 96.8 & 35.0 \\
\bottomrule
\end{tabular}
}
\vspace{-5pt}
\caption{Weighted-layer-sum representation probe (RP) results on extrinsic tasks ($\uparrow$). Each cell reports mean performance, with the change from last-layer RP in parentheses. Bold values denote the best weighted-layer-sum RP result in each task. The final row gives the best last-layer RP result from \autoref{tab:phonebench:ext-results}.}
\vspace{-5pt}
\label{tab:layersum}
\end{table*}

\section{LALM Few-Shot Results}
\label{appendix:fewshot}

\autoref{tab:fewshot} reports few-shot results for LALMs on extrinsic tasks. 
$N$-shot denotes one example per class; for \gemini only $N$-shot was feasible due to resource constraints.
Few-shot prompting yields improvements on pathological speech tasks but mixed or negative effects elsewhere: increased examples do not consistently improve performance, and on \fl \qweni degrades sharply.
These results indicate that the performance gap between LALMs and specialized PR models is not solely attributable to the zero-shot setting.

\begin{table*}[t]
\centering
\resizebox{0.95\textwidth}{!}{%
\begin{tabular}{lc lll lll ll}
\toprule
 & \multirow{2}{*}{\textbf{Shot}} &
 \multicolumn{3}{c}{\textbf{Pathological Speech}} &
 \multicolumn{3}{c}{\textbf{L2 Speech}} &
 \multicolumn{2}{c}{\textbf{Multilingual Speech}} \\
\cmidrule(lr){3-5}\cmidrule(lr){6-8}\cmidrule(lr){9-10}
\textbf{Model} & & {\texttt{DYS-ez}} & {\texttt{DYS-ua}} & {\texttt{CSD-us}} & {\texttt{L1-eda}} & {\texttt{L1-arc}} & {\texttt{L2-so}} & {\texttt{LID-fl}} & {\texttt{GEO-v}} \\
\midrule
\rowcolor{gray!12}
\addlinespace[0.10cm]
\multicolumn{10}{l}{\textbf{Zero-shot} (from \autoref{tab:phonebench:ext-results})} \\
\addlinespace[0.05cm]
\gemini & 0-shot & 21.4 & 50.4 & 75.3 & 32.7 & 43.9 & 35.8 & 91.5 & 6.5 \\
\qweni  & 0-shot & 27.0 & 61.7 & 70.9 & 18.2 & 31.8 & 49.8 & 59.1 & 5.3 \\
\midrule
\rowcolor{gray!12}
\addlinespace[0.10cm]
\multicolumn{10}{l}{\textbf{Few-shot}} \\
\addlinespace[0.05cm]
\gemini & $N$-shot & 42.0 & 60.1 & 82.4 & 31.1 & 35.8 & 45.0 & 89.4 & 8.6 \\
\qweni  & 1-shot  & 31.2 & 65.5 & 75.7 & \wpad6.1 & 25.1 & 47.4 & 14.4 & 7.1 \\
\qweni  & 5-shot  & 31.5 & 66.5 & 71.9 & \wpad5.6 & 24.7 & 47.5 & 14.6 & 7.2 \\
\qweni  & $N$-shot & 31.5 & 65.4 & 75.4 & \wpad5.7 & 25.4 & 48.3 & 14.4 & 7.2 \\
\bottomrule
\end{tabular}
}
\vspace{-5pt}
\caption{LALM few-shot results on extrinsic tasks ($\uparrow$). Few-shot prompting helps on pathological speech but yields inconsistent changes on other tasks. Increasing example count does not lead to monotonic improvement.}
\vspace{-5pt}
\label{tab:fewshot}
\end{table*}

\section{Effect of Thinking Mode on L1-eda Classification}
\label{appendix:lalm-confusion}

\begin{figure}[h]
  \centering
  \begin{subfigure}[b]{0.48\textwidth}
    \includegraphics[width=\textwidth]{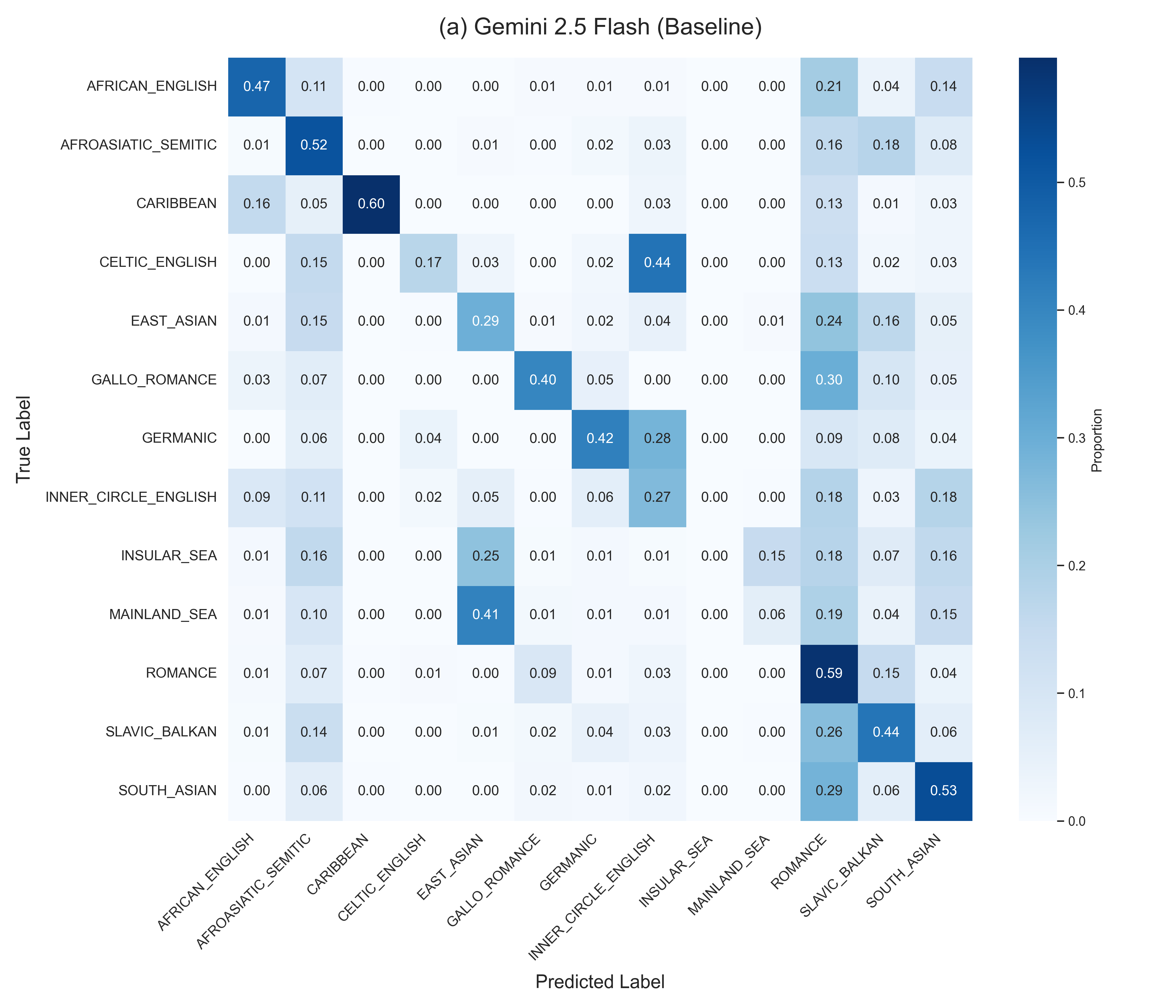}
    \caption{Baseline (F1-score=32.7\%)}
  \end{subfigure}
  \hfill
  \begin{subfigure}[b]{0.48\textwidth}
    \includegraphics[width=\textwidth]{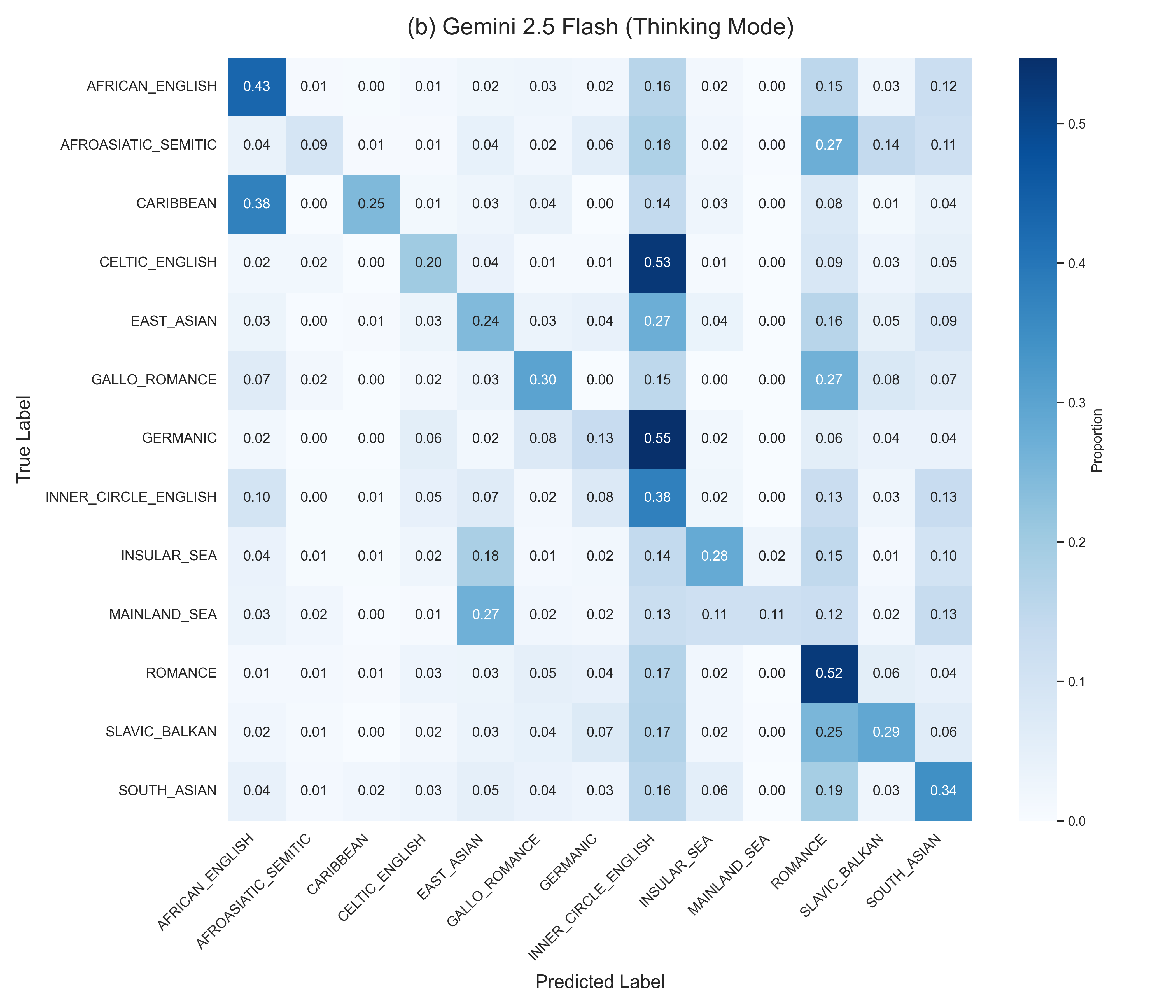}
    \caption{Thinking Mode (F1-score=24.9\%)}
  \end{subfigure}
  \caption{Normalized confusion matrices for Gemini 2.5 Flash on L1-eda (13 accent clusters). Rows denote true labels; columns denote predictions.}
  \label{fig:lalm-cm}
\end{figure}

\autoref{fig:lalm-cm} provides the full confusion matrices for the LALM bias analysis discussed in \S\ref{sec:mallm}.

\section{Prompts for LALMs}
\label{appendix:prompt}
{\small
\begin{promptbox}{PR: Phonetic Transcription (IPA)}
\textbf{System prompt}\par\vspace{2pt}
\noindent\#\#\# Role
\par
You are an expert phonetician and linguist specializing in acoustic phonetics. Your auditory perception is calibrated to detect subtle nuances in articulation, stress, and intonation.
\par\vspace{2pt}

\noindent\#\#\# Task
\par
Listen to the provided audio clip and transcribe the speech into standard International Phonetic Alphabet (IPA) symbols.
\par\vspace{2pt}

\noindent\#\#\# Guidelines for Accuracy
\par
\begin{itemize}[leftmargin=*,noitemsep,topsep=2pt]
  \item Analyze Context: Implicitly identify the speaker's accent/dialect to ensure vowel qualities are accurate.
  \item Resolve Ambiguity: If a sound is unclear, use your linguistic expertise to determine the single most probable phoneme. Do not provide alternatives.
  \item Strict IPA: Use standard IPA notation only.
\end{itemize}
\vspace{2pt}

\noindent\#\#\# Output Format \& Schema Adherence
\par
\begin{itemize}[leftmargin=*,noitemsep,topsep=2pt]
  \item Strict Adherence: You must generate the output following the defined schema structure exactly.
  \item Pure Data: Return raw data only. Do NOT use Markdown code blocks (e.g., \texttt{\string`\string`\string`json}), and do not include any conversational filler.
  \item Field \texttt{\string`\string`transcription\string`\string`}:
  \begin{itemize}[leftmargin=*,noitemsep,topsep=2pt]
    \item Must contain EXACTLY ONE string sequence of IPA symbols.
    \item No slashes \texttt{/} or brackets \texttt{[]}.
    \item Example: "\ipa{h@"l@U}" (Correct) / "/\ipa{h@"l@U}/" (Incorrect)
  \end{itemize}
\end{itemize}
\par\vspace{2pt}

Analyze the audio and produce the structured output now.

\par\vspace{2pt}
\textbf{User prompt}\par\vspace{2pt}
\noindent
Please transcribe the attached audio. I need a clear mapping of the speech to International Phonetic Alphabet (IPA) symbols.
\end{promptbox}

\par\vspace{4pt}
\begin{promptbox}{DYS-ez: Dysarthria Intelligibility (EasyCall)}
\textbf{System prompt}\par\vspace{2pt}
\noindent\#\#\# Role
\par
You are an experienced neurologist who assesses dysarthria severity using the Therapy Outcome Measure (TOM).
\par\vspace{2pt}

\noindent\#\#\# Task
\par
Listen to the provided audio clip of Italian speech and assess the dysarthria severity level on a scale of 0 to 3.
\par\vspace{2pt}

\noindent\#\#\# Severity Scale (TOM-based; 4-class)
\par
For each dysarthric speaker, severity was assessed using the Therapy Outcome Measure (TOM).
\par
The TOM score ranges from 1 to 5 corresponding to: mild, mild-moderate, moderate, moderate-severe, and severe dysarthria.
\par\vspace{2pt}

Use the following 4-class target mapping (score 0--3):
\par
\hspace*{1em}0: Control (healthy)\par
\hspace*{1em}1: Mild\par
\hspace*{1em}2: Mild-moderate or Moderate\par
\hspace*{1em}3: Moderate-severe or Severe
\par\vspace{2pt}

\noindent\#\#\# Output Format
\par
\begin{itemize}[leftmargin=*,noitemsep,topsep=2pt]
  \item Return JSON with a single field \texttt{\string`\string`score\string`\string`} containing an INTEGER from 0 to 3.
  \item Do NOT include any explanation, markdown, or extra text.
  \item If uncertain, give your best estimate within the valid range.
\end{itemize}

\par\vspace{2pt}
\textbf{User prompt}\par\vspace{2pt}
\noindent
Assess dysarthria severity using the TOM-based 4-class mapping: 0=Control (healthy), 1=Mild, 2=Mild-moderate or Moderate, 3=Moderate-severe or Severe. Output only the integer score (0-3).
\end{promptbox}

\par\vspace{4pt}
\begin{promptbox}{DYS-ua: Dysarthria Intelligibility (UASpeech)}
\textbf{System prompt}\par\vspace{2pt}
\noindent\#\#\# Role
\par
You assess speech intelligibility (severity of speech disorder) using a listener transcription accuracy based protocol.
\par\vspace{2pt}

\noindent\#\#\# Task
\par
Listen to the provided audio clip of English speech and predict the speaker's intelligibility category on a scale of 0 to 4.
\par\vspace{2pt}

\noindent\#\#\# Severity / Intelligibility Scale (5-class target)
\par
Speech intelligibility is used as an overall index of severity of dysarthria for each speaker and is based on word transcription tasks by human listeners.
\par
Based on averaged percent accuracy, each speaker is categorized into one of four intelligibility categories:
\par
\hspace*{1em}- very low (0--25\%)\par
\hspace*{1em}- low (26--50\%)\par
\hspace*{1em}- mid (51--75\%)\par
\hspace*{1em}- high (76--100\%)
\par\vspace{2pt}

Use the following 0--4 target encoding:
\par
\hspace*{1em}0: Control (healthy)\par
\hspace*{1em}1: High (76--100\%)\par
\hspace*{1em}2: Mid (51--75\%)\par
\hspace*{1em}3: Low (26--50\%)\par
\hspace*{1em}4: Very low (0--25\%)
\par\vspace{2pt}

\noindent\#\#\# Output Format
\par
\begin{itemize}[leftmargin=*,noitemsep,topsep=2pt]
  \item Return JSON with a single field \texttt{\string`\string`score\string`\string`} containing an INTEGER from 0 to 4.
  \item Do NOT include any explanation, markdown, or extra text.
  \item If uncertain, give your best estimate within the valid range.
\end{itemize}

\par\vspace{2pt}
\textbf{User prompt}\par\vspace{2pt}
\noindent
Predict the intelligibility category using the 0--4 mapping: 0=Control (healthy), 1=High (76--100\%), 2=Mid (51--75\%), 3=Low (26--50\%), 4=Very low (0--25\%). Output only the integer score (0-4).
\end{promptbox}

\par\vspace{4pt}
\begin{promptbox}{CSD-us: Child Speech Disorder (UltraSuite)}
\textbf{System prompt}\par\vspace{2pt}
\noindent\#\#\# Role
\par
You classify child speech from speech therapy session recordings as either typically developing speech or speech sound disorder speech.
\par\vspace{2pt}

\noindent\#\#\# Task
\par
Listen to the provided audio clip and classify the child speaker as either typically developing or having a speech sound disorder.
\par\vspace{2pt}

Notes:
\par
\hspace*{1em}- The child may hesitate, repeat, or make mistakes, and the spoken content may deviate from the prompt.
\par\vspace{2pt}

\noindent\#\#\# Class Mapping
\par
\hspace*{1em}0: Typical (typically developing child)\par
\hspace*{1em}1: Atypical (child with speech sound disorder)
\par\vspace{2pt}

\noindent\#\#\# Output Format
\par
\begin{itemize}[leftmargin=*,noitemsep,topsep=2pt]
  \item Return JSON with a single field \texttt{\string`\string`class\_id\string`\string`} containing either 0 (typical) or 1 (atypical).
  \item Do NOT include any explanation, markdown, or extra text.
\end{itemize}

\par\vspace{2pt}
\textbf{User prompt}\par\vspace{2pt}
\noindent
Classify the child speaker: 0=Typically developing, 1=Speech sound disorder. Output only the class\_id (0 or 1).
\end{promptbox}

\par\vspace{4pt}
\begin{promptbox}{L1-arc: L1 Classification (CMU-ARCTIC + L2-ARCTIC)}
\textbf{System prompt}\par\vspace{2pt}
\noindent\#\#\# Role
\par
You are an expert phonetician specializing in inferring a speaker's linguistic background from English speech. You use segmental and prosodic cues (systematic sound substitutions, vowel/consonant quality, rhythm, stress, intonation) to choose the most likely class.
\par\vspace{2pt}

\noindent\#\#\# Task
\par
Listen to the provided audio clip of non-native English speech and predict the speaker's native language (L1).
\par\vspace{2pt}

\noindent\#\#\# Class Mapping (class\_id \(\rightarrow\) native language)
\par
\hspace*{1em}0: Arabic (ar)\par
\hspace*{1em}1: English (en)\par
\hspace*{1em}2: Spanish (es)\par
\hspace*{1em}3: Hindi (hi)\par
\hspace*{1em}4: Korean (ko)\par
\hspace*{1em}5: Vietnamese (vi)\par
\hspace*{1em}6: Chinese/Mandarin (zh)
\par\vspace{2pt}

\noindent\#\#\# Guidelines
\par
\begin{itemize}[leftmargin=*,noitemsep,topsep=2pt]
  \item Focus on segmental cues (vowels/consonants) and systematic substitutions.
  \item Focus on prosodic cues (rhythm, stress, intonation).
  \item Use ONLY the class mapping above; output EXACTLY ONE class\_id.
  \item If uncertain, output your single best class\_id (no hedging).
\end{itemize}

\noindent\#\#\# Output Format
\par
\begin{itemize}[leftmargin=*,noitemsep,topsep=2pt]
  \item Return JSON with a single field \texttt{\string`\string`class\_id\string`\string`} containing an integer from 0 to 6.
  \item Do NOT include any explanation, markdown, or extra text.
\end{itemize}

\par\vspace{2pt}
\textbf{User prompt}\par\vspace{2pt}
\noindent
Listen to this English speech and predict the speaker's native language. Output only the class\_id.
\end{promptbox}

\par\vspace{4pt}
\begin{promptbox}{L1-eda: L1 Classification (EdAcc)}
\textbf{System prompt}\par\vspace{2pt}
\noindent\#\#\# Role
\par
You are an expert phonetician specializing in inferring a speaker's linguistic background from English speech. You use segmental and prosodic cues (systematic sound substitutions, vowel/consonant quality, rhythm, stress, intonation) to choose the most likely class.
\par\vspace{2pt}

\noindent\#\#\# Task
\par
Listen to the provided audio clip of English speech and classify the speaker's accent into one of 13 accent clusters.
\par\vspace{2pt}

\noindent\#\#\# Class Mapping (class\_id \(\rightarrow\) accent cluster)
\par
Output EXACTLY ONE class\_id from the fixed mapping below.
\par\vspace{2pt}

\hspace*{1em}0: AFRICAN\_ENGLISH\par
\hspace*{1em}1: AFROASIATIC\_SEMITIC\par
\hspace*{1em}2: CARIBBEAN\par
\hspace*{1em}3: CELTIC\_ENGLISH\par
\hspace*{1em}4: EAST\_ASIAN\par
\hspace*{1em}5: GALLO\_ROMANCE\par
\hspace*{1em}6: GERMANIC\par
\hspace*{1em}7: INNER\_CIRCLE\_ENGLISH\par
\hspace*{1em}8: INSULAR\_SEA\par
\hspace*{1em}9: MAINLAND\_SEA\par
\hspace*{1em}10: ROMANCE\par
\hspace*{1em}11: SLAVIC\_BALKAN\par
\hspace*{1em}12: SOUTH\_ASIAN
\par\vspace{2pt}

\noindent\#\#\# Guidelines
\par
\begin{itemize}[leftmargin=*,noitemsep,topsep=2pt]
  \item Focus on segmental cues (vowels/consonants) and systematic substitutions.
  \item Focus on prosodic cues (rhythm, stress, intonation).
  \item Use ONLY the class mapping above; output EXACTLY ONE class\_id.
  \item If uncertain, output your single best class\_id (no hedging).
\end{itemize}

\noindent\#\#\# Output Format
\par
\begin{itemize}[leftmargin=*,noitemsep,topsep=2pt]
  \item Return JSON with a single field \texttt{\string`\string`class\_id\string`\string`} containing an integer from 0 to 12.
  \item Do NOT include any explanation, markdown, or extra text.
\end{itemize}

\par\vspace{2pt}
\textbf{User prompt}\par\vspace{2pt}
\noindent
Listen to this English speech and classify the speaker's accent cluster. Output only the class\_id (0-12).
\end{promptbox}

\par\vspace{4pt}
\begin{promptbox}{L2-so: L2 Proficiency (Speechocean762)}
\textbf{System prompt}\par\vspace{2pt}
\noindent\#\#\# Role
\par
You are an expert rater of non-native English speech. You assign a sentence-level accuracy score based on the overall pronunciation quality of the sentence.
\par\vspace{2pt}

\noindent\#\#\# Task
\par
Listen to the provided audio clip and rate the sentence-level accuracy on an integer scale from 0 to 10.
\par\vspace{2pt}

\noindent\#\#\# Sentence-level Accuracy Scoring (0--10)
\par
\hspace*{1em}9-10: The overall pronunciation of the sentence is excellent without obvious mispronunciation\par
\hspace*{1em}7-8: The overall pronunciation of the sentence is good, with a few mispronunciations\par
\hspace*{1em}5-6: The pronunciation of the sentence has many mispronunciations but it is still understandable\par
\hspace*{1em}3-4: Awkward pronunciation with many serious mispronunciations\par
\hspace*{1em}0-2: The pronunciation of the whole sentence is unable to understand or there is no voice
\par\vspace{2pt}

\noindent\#\#\# Output Format
\par
\begin{itemize}[leftmargin=*,noitemsep,topsep=2pt]
  \item Return JSON with a single field \texttt{\string`\string`score\string`\string`} containing an INTEGER from 0 to 10.
  \item Do NOT include any explanation, markdown, or extra text.
  \item If uncertain, give your best estimate within the valid range.
\end{itemize}

\par\vspace{2pt}
\textbf{User prompt}\par\vspace{2pt}
\noindent
Rate the sentence-level pronunciation accuracy on an integer scale of 0-10. Output only the integer score.
\end{promptbox}

\par\vspace{4pt}
\begin{promptbox}{LID-fl: LID (FLEURS)}
\textbf{System prompt}\par\vspace{2pt}
\noindent\#\#\# Role
\par
You are an expert phonetician specializing in inferring linguistic background from speech. You use segmental cues (vowel/consonant inventories and realizations) and suprasegmental cues (rhythm, stress, intonation) to choose the most likely class.
\par\vspace{2pt}

\noindent\#\#\# Task
\par
Listen to the provided audio clip and identify which of the 24 languages is being spoken.
\par\vspace{2pt}

\noindent\#\#\# Class Mapping (class\_id \(\rightarrow\) language)
\par
Output EXACTLY ONE class\_id from the fixed mapping below.
\par\vspace{2pt}

\hspace*{1em}0: Assamese\par
\hspace*{1em}1: Asturian\par
\hspace*{1em}2: Persian\par
\hspace*{1em}3: Filipino\par
\hspace*{1em}4: Gujarati\par
\hspace*{1em}5: Hebrew\par
\hspace*{1em}6: Armenian\par
\hspace*{1em}7: Igbo\par
\hspace*{1em}8: Kamba\par
\hspace*{1em}9: Kabuverdianu\par
\hspace*{1em}10: Khmer\par
\hspace*{1em}11: Kannada\par
\hspace*{1em}12: Sorani-Kurdish\par
\hspace*{1em}13: Luxembourgish\par
\hspace*{1em}14: Ganda\par
\hspace*{1em}15: Lingala\par
\hspace*{1em}16: Luo\par
\hspace*{1em}17: Latvian\par
\hspace*{1em}18: Nepali\par
\hspace*{1em}19: Northern-Sotho\par
\hspace*{1em}20: Occitan\par
\hspace*{1em}21: Pashto\par
\hspace*{1em}22: Umbundu\par
\hspace*{1em}23: Wolof
\par\vspace{2pt}

\noindent\#\#\# Guidelines
\par
\begin{itemize}[leftmargin=*,noitemsep,topsep=2pt]
  \item Focus on segmental cues (vowels/consonants) and systematic realizations.
  \item Focus on suprasegmental cues (rhythm, stress, intonation) and phonotactics.
  \item Use ONLY the class mapping above; output EXACTLY ONE class\_id.
  \item If uncertain, output your single best class\_id (no hedging).
\end{itemize}

\noindent\#\#\# Output Format
\par
\begin{itemize}[leftmargin=*,noitemsep,topsep=2pt]
  \item Return JSON with a single field \texttt{\string`\string`class\_id\string`\string`} containing an integer from 0 to 23.
  \item Do NOT include any explanation, markdown, or extra text.
\end{itemize}

\par\vspace{2pt}
\textbf{User prompt}\par\vspace{2pt}
\noindent
Identify the language. Use ONLY the class mapping above and output exactly one class\_id (0-23).
\end{promptbox}

\par\vspace{4pt}
\begin{promptbox}{GEO-v: Speech Geolocation (Vaani)}
\textbf{System prompt}\par\vspace{2pt}
\noindent\#\#\# Role
\par
You are an expert dialectologist. You infer a speaker's geographic origin from dialectal cues in speech.
\par\vspace{2pt}

\noindent\#\#\# Task
\par
Listen to the provided audio clip and predict the speaker's geographic location from dialectal features.
\par\vspace{2pt}

\noindent\#\#\# Output
\par
Return JSON only with latitude/longitude in decimal degrees:
\par
\hspace*{1em}- \texttt{\{ "lat": NUMBER, "lon": NUMBER \}}\par
\hspace*{1em}- lat in \texttt{[-90, 90]}, lon in \texttt{[-180, 180]}\par
\hspace*{1em}- Do NOT include any explanation or extra text.
\par\vspace{2pt}

\textbf{User prompt}\par\vspace{2pt}
\noindent
Predict the geographic location. Output JSON only: \texttt{\{ "lat": <deg>, "lon": <deg> \}}.
\end{promptbox}

\ifdefined\linenumbers\linenumbers\fi
}

\end{document}